\documentclass[10pt,onecolumn,letterpaper]{article}

\usepackage[top=1.5in,bottom=1.5in,left=1.2in,right=1.2in,columnsep=0.35in]{geometry}

\usepackage{indentfirst}
\usepackage{xcolor}
\usepackage{hyperref}
\usepackage{enumitem}
\usepackage{setspace}
\usepackage{algorithm}
\usepackage{algorithmic}
\usepackage{amsmath}
\usepackage{amsthm}
\usepackage{amssymb}
\usepackage{natbib}
\usepackage{stmaryrd}
\usepackage{hyperref}
\usepackage{tabularx}
\usepackage{booktabs} 
\usepackage{color}
\usepackage{graphicx}
\usepackage{caption}

\usepackage{subfigure}
\usepackage{pifont}

\usepackage{authblk}

\newtheorem{lemma}{Lemma}[section]
\newtheorem{theorem}{Theorem}[section]

\newtheorem{assumption}{Assumption}        
\theoremstyle{definition}
\newtheorem{definition}{Definition}
\newtheorem{remark}{Remark}[section]

\author[1]{Wenjie Li}
\author[1]{Qifan Song}
\author[2]{Jean Honorio}
\author[1,3,4]{Guang Lin}
\affil[1]{Department of Statistics, Purdue University}
\affil[2]{Department of Computer Science, Purdue University}
\affil[3]{Department of Mathematics, Statistics, Purdue University}
\affil[4]{School of Mechanical Engineering, Purdue University }
\date{}

\begin{document}

\title{Federated $\mathcal{X}$-Armed Bandit}
\maketitle

\begin{abstract}
This work establishes the first framework of federated $\mathcal{X}$-armed bandit, where different clients face heterogeneous local objective functions defined on the same domain and are required to collaboratively figure out the global optimum. We propose the first federated algorithm for such problems, named \texttt{Fed-PNE}. By utilizing the topological structure of the global objective inside the hierarchical partitioning and the weak smoothness property, our algorithm achieves sublinear cumulative regret with respect to both the number of clients and the evaluation budget. Meanwhile, it only requires logarithmic communications between the central server and clients, protecting the client privacy. Experimental results on synthetic functions and real datasets validate the advantages of \texttt{Fed-PNE} over various centralized and federated baseline algorithms.
\end{abstract}

\newenvironment{list1}{
  \begin{list}{$\bullet$}{%
      \setlength{\itemsep}{3pt}
      \setlength{\parsep}{5pt} \setlength{\parskip}{0in}
      \setlength{\topsep}{5pt} \setlength{\partopsep}{0in}
      \setlength{\leftmargin}{15pt}}}{\end{list}}

\section{Introduction}
\label{sec: introduction}

Federated bandit is a newly-developed bandit problem that incorporates federated learning with sequential decision making \citep{mcmahan2017communication, shi2021federateda}.
Unlike the traditional bandit models where the exploration-exploitation tradeoff is the only major concern, federated bandit problem also takes account of the modern concerns of data heterogeneity and privacy protection towards trustworthy machine learning.  In particular, in the federated learning paradigm, the data available to each client could be drawn from non-i.i.d distributions, making collaborations between the clients necessary to make valid inferences for the aggregated global model. However, due to user privacy concerns and the large communication cost, such collaborations across the clients must be restricted and avoid direct transmissions of the local data. To make correct decisions in the future, the clients have to utilize the limited communications from each other and coordinate exploration and exploitation correspondingly. 

To the best of our knowledge, existing results of federated bandits, such as \citet{dubey2020differentially, huang2021federated, shi2021federateda, shi2021federatedb}, focus on either the case where the number of arms is finite (multi-armed bandit), or the case where the expected reward is a linear function of the chosen arm (linear contextual bandit). However, for problems such as dynamic pricing \citep{chen2022APrimal} and hyper-parameter optimization \citep{shang2019general}, the available arms are often defined on a domain $\mathcal{X}$ with infinite or even uncountable cardinality, and the reward function is usually non-linear with respect to the metric employed by the domain $\mathcal{X}$. These problems challenge the applications of existing federated bandit algorithms to more complicated real-world problems. Two applications (Figure \ref{fig: example}) that motivate our study of federated $\mathcal{X}$-armed bandit are given below.

\newcommand{\cmark}{\ding{51}}%
\newcommand{\xmark}{\ding{55}}%

\begin{table*}

\caption{\footnotesize Comparison of the (client-wise) regret upper bounds, the communication cost for sufficiently large $T$ and the other properties. \textbf{Columns}: ``Commun. cost'' refer to communication cost.
``Conf." refers to whether the raw rewards of one client are kept confidential from the other clients and only statistical summary is shared. ``Heterogeneity" refers to whether the client functions are different/the same. \textbf{Rows:} \texttt{HCT} is a single-client $\mathcal{X}$-armed bandit algorithm. \texttt{BLiN} is a batched-$\mathcal{X}$-armed bandit algorithm. \texttt{Centralized} results are adapted from the centralized algorithms such as \texttt{HOO} \citep{bubeck2011X} and \texttt{HCT} \citep{azar2014online} by assuming that the server makes all the decisions with access to all client-wise information. \textbf{Notation:} $M$ denotes the number of clients; $T$ denotes the budget (time horizon) and $d$ denotes the near-optimality dimension in Assumption \ref{assumption: near-optimality dimension}.   }
\label{tab: regret_compare}
\centering
\begin{tabular}{lllc c}
\hline Bandit algorithms & Average Regret  & Commun.cost & Conf. & Heterogeneity\\
\hline  
\texttt{HCT} 
 & $\widetilde{\mathcal{O}}\left( T^{\frac{d+1}{d+2}}\right)$ & N.A. & N.A. & \xmark\\
 
\texttt{BLiN}
 & $ \widetilde{\mathcal{O}} \left(  T^{\frac{d+1}{d+2}} \right)$ & N.A. & \cmark & \xmark \\

\texttt{Centralized}
 & $\widetilde{\mathcal{O}} \left( M^{-\frac{1}{d+2}}  T^{\frac{d+1}{d+2}} \right)$ & ${\mathcal{O}}(MT)$ & \xmark & \cmark \\

This work & $\widetilde{\mathcal{O}} \left( M^{-\frac{1}{d+2}}  T^{\frac{d+1}{d+2}} \right)$ & $\widetilde{\mathcal{O}}(M \log T\vee MT^{\frac{d}{d+2}})$ & \cmark &\cmark\\

 



\hline 
\vspace{-30pt}
\end{tabular}
\end{table*}

\begin{list1}
    \item \textbf{Federated medicine dosage recommendation.} For the dosage recommendation of a newly-invented medicine/vaccine (in terms of volume or weight), the clinical trials could be conducted at multiple hospitals (clients). To protect patients' privacy, hospitals cannot directly share the treatment result of each trial (reward). Moreover, because of the demographic difference among the patient groups, the best dosage obtained at each hospital (i.e., the optimal of local objectives) could be different from the optimal recommended dosage for entire population of the state (i.e., the optimal of the global objective). Researchers needs to collaboratively find the global optimal dosage by exploring and exploiting the local data.
    
    \item \textbf{Federated hyper-parameter optimization. }
    An important application of automating machine learning workflows with minimal human intervention consists of
    hyper-parameter optimization for ML models, e.g., learning rate, neural network architecture, etc. Many modern data are collected by mobile devices (clients).
    The model performance (reward) of each hyper-parameter setting could be different for each mobile device (i.e., local objectives) due to user heterogeneity. To fully utilize the whole dataset (i.e., global objective) for hyperparameter optimization such that the obtained auto-ML model can work seamlessly for diverse scenarios, the central server need to coordinate the local search properly without violating the regulations of consumer data privacy.

\end{list1}

In the aforementioned examples, the reward objectives are defined on a domain $\mathcal{X}$, which can often be formatted as a region of $\mathbb{R}^d$ and has infinite cardinality. Moreover, the objectives (both local and global ones) are highly nonlinear mapping with respect to the arm chosen due to the complex nature of the problem. Therefore, the basic assumptions of federated multi-armed bandit or federated linear contextual bandit algorithms are violated, and thus the existing federated bandit algorithms cannot apply or   perform well on such problems.

Under the classical setting where centralized data is immediately available, $\mathcal{X}$-armed bandit algorithms such as \texttt{HOO} and \texttt{HCT} have been proposed to find the optimal arm inside the domain $\mathcal{X}$ \citep{bubeck2011X, azar2014online}. However, these algorithms cannot be trivially adapted to the task of finding the global optimum when there are multiple clients and  limit communications. The local objectives could have very different landscapes across the clients due to the non-i.i.d local datasets, and no efficient communication method has been established between $\mathcal{X}$-armed bandit algorithms that run on the local data sets.  In this work, we propose a new federated algorithm where all the clients collaboratively learn the best solution to the global $\mathcal{X}$-armed bandit model on average, while few communications (in terms of the amount and the frequency) are required so that the privacy of each client is preserved.

We highlight our major contributions as follows. 
\begin{list1}
    \item \textbf{Federated $\mathcal{X}$-armed bandit.} We establish the first framework of the federated $\mathcal{X}$-armed bandit problem (Sec. \ref{sec: prelim}), which naturally connects the $\mathcal{X}$-armed bandit problem with the characteristics of federated learning. The new framework introduces many new challenges to $\mathcal{X}$-armed bandit including (1) potential severe heterogeneity among the local objectives due to non-i.i.d local data sets, (2) the non-accessibility of the global objective for all local clients or the central server, and (3) the restriction of communications between the server and the clients.
    
    \item \textbf{New algorithm with desirable regret.} We propose a new algorithm for the federated $\mathcal{X}$-armed bandit problem named \texttt{Fed-PNE}. Inspired by the heuristic of arm elimination in multi-armed bandits \citep{lattimore2020bandit}, the new algorithm performs \textit{hierarchical node elimination} in the domain $\mathcal{X}$. More importantly, it incorporates efficient communications between the server and the clients to transmit information while protecting client-privacy. We establish the sublinear cumulative regret upper bound of the proposed algorithm as well as the bound of the communication cost. Theoretically, we prove that \texttt{Fed-PNE} utilizes the advantage of federation and at the same time has high communication efficiency (Sec. \ref{sec: algorithm}). 
    We also provide a regret lower bound analysis to justify the tightness of our upper bound.
    Theoretical comparisons of our regret bounds with existing bounds are provided in Table \ref{tab: regret_compare}.

    \item  \textbf{Empirical results.} By examining the empirical performance of our \texttt{Fed-PNE} algorithm on both synthetic functions and real datasets (Sec \ref{sec: experiments}), we verify the correctness of our theoretical results. We show the advantages of \texttt{Fed-PNE} over centralized $\mathcal{X}$-armed and kernelized bandit algorithm, and federated neural and multi-armed bandit algorithm. The empirical results exhibit the usefulness of our algorithm in real-life applications.
\end{list1}

\begin{figure*}
    \centering
    \includegraphics[width=0.9\textwidth, height=1.8in]{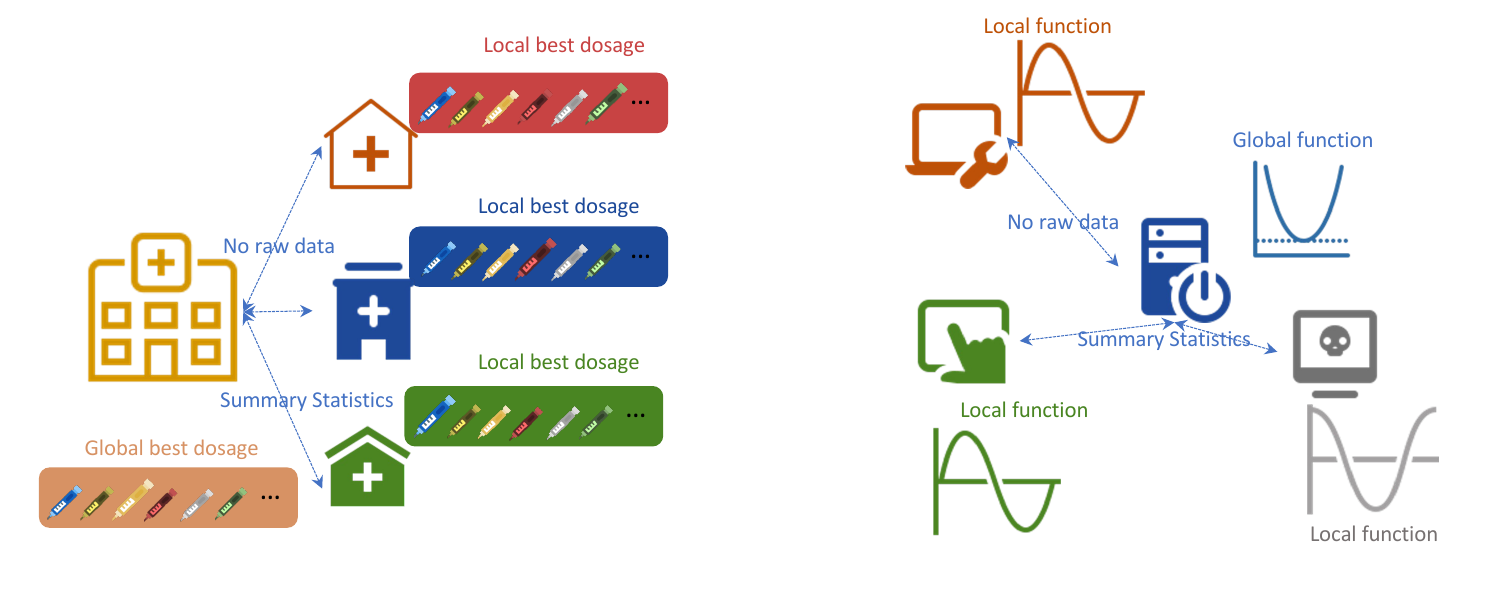}
    \vspace{-15pt}
     \caption{Examples of real-life applications that motivate the federated $\mathcal{X}$-armed bandit problem. Left: federated medicine dosage recommendation. Right: Federated hyper-parameter optimization. }
    \vspace{-10pt}
    \label{fig: example}
\end{figure*}

\section{Preliminaries}
\label{sec: prelim}

We first introduce the preliminary concepts and notations used in this paper. For a real number $a \in \mathbb{R}$, we use $\lceil a \rceil$ to represent the smallest integer larger than $a$. For an integer $N \in \mathbb{N}$, we use $[N]$ to represent the set of integers $\{1, 2,\cdots, N\}$. For a set $A$, $|A|$ denotes the number of elements in $A$. We use $\widetilde{\mathcal{O}}(\cdot)$ to hide the logarithmic terms in big-$\mathcal{O}$ notations, i.e., for two functions $a(n), b(n)$, $a(n) =\widetilde{\mathcal{O}}(b(n)) $ represents that $a(n)/b(n)\leq \log^{k}(n), \forall n >0$ for some $k>0$.

\subsection{Problem Formulation and Performance Measure}
Let $\mathcal{X}$ be a measurable space of arms. We model the problem as a federated $\mathcal{X}$-armed bandit setting where a total of $M$ clients respectively have the access to $M$ different \textit{local objectives} $f_m(x): \mathcal{X} \mapsto \mathbb{R}$, which could be non-convex, non-differentiable and even non-continuous. Given a limited number of rounds $T$, each client $ \in [M]$ chooses a point $x_{m, t} \in \mathcal{X}$ at each round $t \in [T]$ and observes a noisy feedback $r_{m,t} \in [0, 1]$ defined as $ r_{m, t}:= f_m(x_{m, t}) + \epsilon_{m,t}$ , where $\epsilon_{m,t}$ is a zero-mean and bounded random noise independent from previous observations or other clients' observations. The goal of the clients is to find the point that maximizes the \textit{global objective} $f(x)$, which is defined to be the average of the local objectives, i.e.,
\begin{equation}
\nonumber
    f(x) := \frac{1}{M} \sum_{m=1}^M f_m (x).
\end{equation}
However, the global objective is not accessible by any client. The only information that the clients have access to is: (1) noisy evaluations of their own local objective functions $f_m(x)$, and (2) communications between themselves and the central server. For the global objective, we assume that there is at least one global maximizer $x^* \in \mathcal{X}$ such that $f(x^*) = \sup_{x\in \mathcal{X}} f(x) = f^*$. Given the sequence of the points chosen by the clients $\{x_{m,t}\}_{m=1, t=1}^{M,T}$, the performance of the clients is measured by the expectation of the \textit{cumulative regret}, defined as
\begin{equation}
\nonumber
    \mathbb{E}\left[R(T)\right] :=  \mathbb{E}\left[ \sum_{t=1}^T \sum_{m=1}^M \left(f^*  - f(x_{m,t}) \right)\right].  
\end{equation}
Another possible measure of algorithm performance is the so-called \textit{simple regret} which only evaluates the goodness of optimizer in the final round, i.e., 
$r(T) = \sum_{m=1}^M (f^*  - f(x_{m,T}))$. This paper aligns with the standard federated bandit analysis framework and focuses on cumulative regret only \citep{shi2021federateda, huang2021federated}. Moreover, as mentioned by \citet{bubeck2011X}, we always have $\mathbb{E}[r(T)] \leq \mathbb{E}[R(T)]/T$ if we select the path via a cumulative regret-based policy.

\subsection{Hierarchical Partitioning of the Parameter Space}
Following the recent progress in centralized $\mathcal{X}$-armed bandit \citep[e.g.,][]{azar2014online, shang2019general, bartlett2019simple}, we utilize a pre-defined infinitely-deep hierarchical partitioning $\mathcal{P} := \{\mathcal{P}_{h,i}\}_{h,i}$ of the parameter space $\mathcal{X}$ to optimize the objective functions. The hierarchical partition discretizes the space by recursively defining the following relationship:
\begin{equation}
\nonumber
    \mathcal{P}_{0, 1} := \mathcal{X}, \qquad \mathcal{P}_{h,i} := \bigcup_{j=0}^{k-1}  \mathcal{P}_{h+1, ki-j},
\end{equation}
where $k$ is the (maximum) number of disjoint children for one node, and for every node $\mathcal{P}_{h,i}$, $(h,i)$ denotes the depth and the index of the node inside the partition. Each node $\mathcal{P}_{h,i}$ on depth $h$ is partitioned into $k$ children on depth $h+1$, while the union of all the nodes on each depth $h$ equals the parameter set $\mathcal{X}$. The partition is chosen before the optimization process and the same partition of the space $\mathcal{X}$ is shared and used by all the $M$ clients as the partition itself reveals no information of the reward distributions of local objectives. A simple and intuitive example is a binary equal-sized partition on the domain $\mathcal{X} = [0, 1]$, where each node on depth $h$ has length $(0.5)^{h}$.

\subsection{Communication Model and Privacy Concerns.} 

Similar to the setting of federated multi-armed bandit \citep{shi2021federateda, huang2021federated}, we assume that there exists a central server that coordinates the behaviors of all the different clients. The server has access to the same partition of the parameter space used by all the clients, and is able to communicate with the clients. Due to privacy concerns, the client-side algorithm should keep the reward of each evaluation confidential and the only things that can be transmitted to the server are the local statistical summary of the rewards. The clients are not allowed to communicate with each other. In accordance to \citet{mcmahan2017communication, shi2021federateda}, we assume that the server and the clients are fully synchronized. Although the clients can communicate with the server, the number of clients $M$ could be very large and thus the communication would be very costly. We take into account such communication cost in our algorithm design and the theoretical analysis.

\subsection{Definitions and Assumptions}

To analyze the performance of the proposed algorithms, we use the following set of definitions and assumptions, which are also present in the prior works on $\mathcal{X}$-armed bandit \citep{bubeck2011X, azar2014online}.
\begin{assumption}
\label{assumption: dissimilarity}
\textbf{\upshape(Dissimilarity)}
        The space $\mathcal{X}$ is equipped with a dissimilarity function $\ell: \mathcal{X}^2 \mapsto \mathbb{R}$ such that $\ell(x, x')\geq 0, \forall (x, x') \in \mathcal{X}^2$ and $\ell(x, x) = 0$ 
\end{assumption}
Throughout this work, we assume that $\mathcal{X}$ satisfies Assumption \ref{assumption: dissimilarity}. Given the dissimilarity function $\ell$, the diameter of a set $\mathcal{A} \subset \mathcal{X}$ is defined as $\operatorname{diam}\left(\mathcal{A}\right) = \sup_{x, y \in \mathcal{A}} \ell(x, y) $. The open ball of radius $r$ and with center $c$ is then defined as $\mathcal{B}(c, r) = \{x \in \mathcal{X}: \ell(x, c) \leq r\}$. We now introduce the local smoothness assumptions.

\begin{assumption}\textbf{\upshape (Local Smoothness) }
\label{assumption: local_smoothness}
We assume that there exist constants $\nu_1, \nu_2 > 0$, and $0< \rho < 1$ such that for all nodes $\mathcal{P}_{h,i}, \mathcal{P}_{h,j}\in \mathcal{P}$ on depth $h$,
\begin{list1}
    \item $\operatorname{diam}\left(\mathcal{P}_{h, i}\right) \leq \nu_{1} \rho^{h}$
    \item  $\exists x_{h, i}^{\circ} \in \mathcal{P}_{h, i}$ s.t. $\mathcal{B}_{h, i} {:=} \mathcal{B}\left(x_{h, i}^{\circ}, \nu_{2} \rho^{h}\right) \subset \mathcal{P}_{h, i}$
    \item $\mathcal{B}_{h, i} \cap \mathcal{B}_{h, j}=\emptyset$ for all $1 \leq i<j \leq k^{h}$.

    \item The global objective function $f$ satisfies that for all $x, y \in \mathcal{X}$, we have 
    \begin{equation}
    \nonumber
        f^* - f(y) \leq f^*- f(x) + \max \left \{f^*- f(x), \ell(x, y) \right \}
    \end{equation}
\end{list1}
\end{assumption}

\begin{remark}

Similar to the existing works on the $\mathcal{X}$-armed bandit problem, the dissimilarity function $\ell$ is not an explicit input required by our \texttt{Fed-PNE} algorithm and only the smoothness constants $\nu_1, \rho$ are accessed \citep{bubeck2011X, azar2014online}. As mentioned by \citet{bubeck2011X, Grill2015Blackbox}, most regular functions satisfy Assumption \ref{assumption: local_smoothness} on the standard equal-sized partition with accessible $\nu_1$ and $\rho$. 
\end{remark}

Finally, we introduce the definition of the near-optimality dimension, which measures the number of near-optimal regions and thus the difficulty of the problem \citep{azar2014online}.

\begin{assumption}
\label{assumption: near-optimality dimension}
{\upshape \textbf{(Near-optimality dimension)}}
Let $\epsilon = 6 \nu_1 \rho^h$ and $\epsilon' = \rho^h < \epsilon$, for any subset of $\epsilon$-optimal nodes $\mathcal{X}_{\epsilon} = \{ x\in \mathcal{X} : f^* - f(x) \leq \epsilon \}$, there exists a constant $C$ such that $\mathcal{N}(\mathcal{X}_{\epsilon}, \ell, \epsilon') \leq C (\epsilon')^{-d}$, where $d$ is the near-optimality dimension of function $f$ and $\mathcal{N}(\mathcal{X}_{\epsilon}, \ell, \epsilon')$ is the $\epsilon'$-cover number of the set $\mathcal{X}_{\epsilon}$ w.r.t. the dissimilarity $\ell$.
\end{assumption}

\begin{remark}
Some recent progress of solving centralized $\mathcal{X}$-armed bandit problem such as \citet{shang2019general, bartlett2019simple} have proposed an even weaker version of Assumption \ref{assumption: local_smoothness}, i.e., the \textit{local smoothness without a metric} assumption. Correspondingly, they define the complexity measure named \textit{near-optimal dimension w.r.t. the partition $\mathcal{P}$}. 
However, it is highly non-trivial to directly adopt this weaker local smoothness assumption in the federated $\mathcal{X}$-armed bandit problem. The limited communications and the weak assumption will lead to continual sampling in the sub-optimal regions, and thus yielding large cumulative regrets.
As a pioneer work in federated $\mathcal{X}$-armed bandit, we choose to use the slightly stronger assumptions in \citet{bubeck2011X} so that theoretical guarantees of our \texttt{Fed-PNE} algorithm can be successfully established. Weakening our set of assumption while keeping the regret bound guarantee is an interesting future work direction.  
\end{remark}

\section{Algorithm and Analysis}
\label{sec: algorithm}

The federated $\mathcal{X}$-armed bandit problem encounters several challenges, the core of which is to accommodate the heterogeneity among local objectives with limited communications. Hence, how to design an efficient communication pattern and construct an unbiased estimation of the global objective while taking advantage of the large number of clients is a crucial component in algorithmic design. Moreover, since local rewards are not instantaneously observable due to communication limitation, any algorithm that ``uses instant rewords of each time step to estimate the optimal region with high confidence", e.g., \texttt{HOO} \citep{bubeck2011X} and \texttt{HCT} \citep{azar2014online}, cannot be directly applied to this problem. Instead, an algorithm that gradually eliminates the sub-optimal regions in phases is preferred.

In this section,  we propose the new algorithm to solve the above challenges, show its uniqueness compared with prior algorithms, and provide its theoretical analysis.

\subsection{The \texttt{Fed-PNE} Algorithm}

We propose the new Federated-Phased-Node-Elimination (\texttt{Fed-PNE}) algorithm, which consists of one client-side algorithm (Algorithm \ref{alg: client}) and one server-side algorithm (Algorithm \ref{alg: server}). The \texttt{Fed-PNE} algorithm runs in dynamic phases and it utilizes the hierarchical partition to gradually find the optimum by eliminating different regions of the domain. For a node $\mathcal{P}_{h,i} \in \mathcal{P}$, since its depth $h$ and index $i$ uniquely identifies the node, we will use $(h,i)$ to index the nodes in the elimination and expansion process. We use $\mathcal{K}^p$ to denote the indices of active nodes that need to be sampled in phase $p$ and $\mathcal{E}^p$ for the indices of nodes that need to be eliminated. To obtain a reward $r$ over a node $\mathcal{P}_{h,i}$ (i.e., pull a node), the client evaluate the local objective at some $x$ where $x$ is either uniformly sampled from the node as in \citet{bubeck2011X} or some pre-defined point in the node as in \citet{azar2014online}. The regret analysis will only be slightly different because of the smoothness assumption. In Section \ref{subsec: theoretical_analysis}, we have used the latter strategy to derive our regret bound.

\begin{algorithm}
   \caption{ \texttt{Fed-PNE: $m$-th client}}
   \label{alg: client}
\begin{algorithmic}[1]
   \STATE \textbf{Input:} $k$-nary partition $\mathcal{P}$
   \STATE \textbf{Initialize} $p = 0$
   \WHILE{not reaching the time horizon $T$}
    \STATE Update $p = p+1$
    \STATE  Receive  $\{\mathcal{P}_{h, i}, t_{m, h, i}\}_{(h, i) \in \mathcal{K}^p}$ from the server
    \FOR{ $\mathcal{P}_{h, i}$ with $(h, i) \in \mathcal{K}^p$  }
    \STATE Pull the node for $ t_{m, h, i}$ times, receive rewards $\{r_{m, h, i, t}\}_{t=1}^{t_{m, h, i}}$
    \STATE Calculate the local mean estimate $\widehat{\mu}_{m, h,i} = \frac{1}{ t_{m, h, i}} \sum_{t} r_{m, h, i, t}$
    \ENDFOR
    \STATE Send the local estimates $\{\widehat{\mu}_{m, h,i}\}_{(h,i) \in \mathcal{K}^p}$ to the server
   \ENDWHILE
\end{algorithmic}
\end{algorithm}

\textbf{Algorithm Explanation}: At initialization, the server starts from the root of the partition $\mathcal{K}^1 = \{(0, 1)\}$. At the beginning of each phase $p>0$, the server expands the exploration tree as described in Algorithm \ref{alg: server} and the set $\mathcal{K}^p$ until the criterion $|\mathcal{K}^p| \tau_{h}\geq M$ is satisfied, where the threshold number $\tau_h$ is the minimum required number of times each node on depth $h$ needs to be pulled, defined as
%
$\tau_h := \left\lceil \frac{c^2 \log(c_1T/\delta)}{  \nu_1^2}  \rho^{-2h} \right \rceil$
where $c, c_1$ are two absolute constants, and $\delta$ is the confidence (details in Lemma \ref{lem: good_event}). The number of times  $t_{m, h, i}$ each node $\mathcal{P}_{h,i}$ has to be sampled by each client $m$ and the phase length $|\mathcal{T}^p|$ are then computed. This unique expansion criteria and sampling scheme guarantee four important things at the same time: (1) Every client samples every node at least one time so that the global objective is explored; (2) The empirical averages in line 12 of Algorithm \ref{alg: server} are unbiased estimators of the global function values for every node; (3) Every node is sampled enough number of times (larger than $\tau_h$); (4) The waste of budget due to the limitation on communication is minimized. After the broadcast in line 9, every client receives $\{\mathcal{P}_{h, i}, t_{m, h, i}\}_{(h, i) \in \mathcal{K}^p}$ from the server. 

Next, the clients perform the exploration and send only the empirical reward averages $\widehat{\mu}_{m, h,i}$ back to the server, as in Algorithm \ref{alg: client}. The server then computes the best node, denoted by $\mathcal{P}_{h^p,i^p} $, and decides the elimination set $\mathcal{E}^p$ by the following selection criteria.
\begin{equation}\label{eqn: elimination_set}
\begin{split}
   \mathcal{E}^p := \{(h, i) \in \mathcal{K}^p \mid \widehat{\mu}_{h,i} &+ b_{h,i} + \nu_1\rho^h < \widehat{\mu}_{h^p,i^p} - b_{h^p,i^p}  \}
     \end{split}
\end{equation}
where $ b_{h,i} = c \sqrt{{\log(c_1 T/\delta)}/{T_{h,i}}}$ and $T_{h, i} = M t_{m, h, i}$. In other words, for any node $\mathcal{P}_{h,i}$ such that $(h,i) \in \mathcal{E}^p$, the function value of the global objective inside the node is much worse than the function value in the best node with high probability, and thus can be safely eliminated. The server then eliminate the bad nodes and proceed to the next phase with the new set $\mathcal{K}^{p+1}$, which consists of nodes that are children of un-eliminated nodes in the previous phase, as shown in line 15-16 in Algorithm \ref{alg: server}.

\begin{algorithm}
   \caption{ \texttt{Fed-PNE: server}}
   \label{alg: server}
\begin{algorithmic}[1]
   \STATE \textbf{Input:} $k$-nary partition $\mathcal{P}$, smooth parameters $\nu_1, \rho$
   \STATE \textbf{Initialize} $ \mathcal{K}^1 = \{(0, 1)\}, h = 0, p = 0$
   \WHILE{not reaching the time horizon $T$}
   \STATE $p = p + 1; h = h+1$
    \WHILE{$|\mathcal{K}^p| \tau_{h}\leq M$ or $\tau_h \leq 1$}
    \STATE $\mathcal{K}^{p} = \left\{(h'+1, ki-j) \mid \forall (h',i) \in \mathcal{K}^p, j \in [k-1] \right\}$ \\  Renew  $h = h+1$ 
    \ENDWHILE
    \STATE Compute the number  $ t_{m,h,i} =  \left \lceil  \frac{\tau_{h}}{M} \right \rceil $ and the phase length $|\mathcal{T}^p| = |\mathcal{K}^p| t_{m, h, i}$ 
    \STATE  Broadcast the set of nodes and pulled times $\{\mathcal{P}_{h, i}, t_{m, h, i}\}_{(h, i) \in \mathcal{K}^p}$ to every client $m$
    \STATE  Receive local estimates $\{\widehat{\mu}_{m, h,i}\}_{m \in [M], (h,i) \in \mathcal{K}^p}$ from the clients
    \FOR{ every $(h, i) \in \mathcal{K}^p$  }
    \STATE Calculate the global mean estimate $\widehat{\mu}_{h,i} = \frac{1}{M} \sum_{m=1}^M \widehat{\mu}_{m, h,i}$
    \ENDFOR
    \STATE Compute $(h^p,i^p) = \arg\max_{(h, i) \in \mathcal{K}^p}\widehat{\mu}_{h,i} $
    \STATE Compute the elimination set $\mathcal{E}^p = \left \{(h, i) \in \mathcal{K}^p \mid \widehat{\mu}_{h,i} + b_{h,i} + \nu_1\rho^h < \widehat{\mu}_{h^p,i^p} - b_{h^p,i^p}  \right \}$  
    \STATE Compute the new set of nodes $ \mathcal{K}^{p+1} = \left\{(h+1,ki-j) \mid (h,i) \in (\mathcal{K}^p \setminus \mathcal{E}^p), j \in [k-1] \right\}$
   \ENDWHILE
\end{algorithmic}
\end{algorithm}

\begin{remark}
\texttt{Fed-PNE} is very different from the traditional Phased-Elimination (\texttt{PE}) algorithm in multi-armed bandit\citep{lattimore2020bandit}, though both algorithms utilize the idea of successive elimination of the suboptimal arms/nodes. Apart from the obvious uniqueness in the algorithm design such as line 5-8, 15-16 in Algorithm \ref{alg: server}, \texttt{Fed-PNE} also introduces the new idea of ``node elimination", which is based on the hierarchical partitioning of the parameter space. Even if we treat nodes in the partition as the ``arms" in multi-armed bandit, \texttt{Fed-PNE} is still unique in the following aspects:

\begin{list1}
    \item \texttt{Fed-PNE} utilizes the hierarchical partition and gradually eliminate nodes on deeper layers that represent smaller and smaller regions in domain $\mathcal{X}$. The nodes can not be eliminated until the algorithm reaches their layer in the partition. In other words, the eliminated nodes are different in nature, whereas in multi-armed bandit problem, the arms have equal roles and can be eliminated in any phase;
    
    \item While eliminating the sub-optimal regions, \texttt{Fed-PNE} also explores deeper in the partition and splits one node into multiple nodes, which means that the number of nodes to be sampled may increase instead of decrease as $p$ increases. However, the number of remaining arms never increases in \texttt{PE}. This feature also brings more difficulty to the analysis of \texttt{FedPNE} because the phase length is dynamic instead of fixed;
    
    \item The elimination criteria in Eqn. \eqref{eqn: elimination_set} is carefully designed so that non-optimal nodes are gradually eliminated. The design takes account of not only the Upper-Confidence Bound (UCB) terms $b_{h,i}$ for statistical uncertainty, but also the smoothness term $\nu_1\rho^h$, which reflects for the variation of the objective function inside one node.
\end{list1}
\end{remark}

\begin{remark}
  Compared with centralized $\mathcal{X}$-armed bandit algorithms such as \texttt{HOO} and \texttt{HCT}, our algorithm is also unique in the sense that none of them can deal with the federated, heterogeneous learning setting. The collaboration scheme and the length of each phase $\mathcal{T}^p$ is carefully designed so that the communication to the server is effective. It is worth mentioning that our algorithm requires the parameters $\nu_1, \rho$ as part of the input, which measures how fast the diameter of a node shrinks in the partition. These parameters are important because they characterize the smoothness of the global objective and we need them to determine the threshold $\tau_h$ and the elimination set $\mathcal{E}^p$. This information is crucial to ensure that validity of cumulative regret analysis theorems even for centralized $\mathcal{X}$-armed bandit problems. Most of existing $\mathcal{X}$-armed bandit algorithms, such as \citet{bubeck2011X}, \citet{azar2014online} and \citet{li2021optimumstatistical}, require these parameters. 
\end{remark}

\subsection{Theoretical Analysis}
\label{subsec: theoretical_analysis}
We provide the upper bound on the expected cumulative regret of the proposed \texttt{Fed-PNE} algorithm as follows, which exhibits our theoretical advantage over non-federated algorithms.

\begin{theorem}
\label{thm: regret_upper_bound}
Suppose that $f(x)$ satisfies Assumption \ref{assumption: local_smoothness}, and $d$ is the near-optimality dimension of the global objective $f$ as defined in Assumption \ref{assumption: near-optimality dimension}. Setting $\delta = 1/M$, we have the following upper bound on the expected cumulative regret of the \texttt{Fed-PNE} algorithm.
\begin{equation}
\begin{aligned}
\nonumber
  \mathbb{E}[R (T)] 
& \leq C_1 M^{1-\frac{1}{2\log_k \rho}}  + C_2 \left( M^{\frac{d+1}{d+2}}  T^{\frac{d+1}{d+2}} (\log (MT))^{\frac{1}{d+2}} \right)\\
\end{aligned}
\end{equation}
where $C_1$ and $C_2$ are two absolute constants that do not depend on $M$ and $T$. Moreover, the number of communication rounds of \texttt{Fed-PNE} scales as $\widetilde{\mathcal{O}}(M \log T)$
\end{theorem}

\begin{remark}
The proof of the above theorem and the exact values of the two constants are relegated to Appendix \ref{app: notaions_and_lemmas}, \ref{app: Main_Proof}. Theorem \ref{thm: regret_upper_bound} displays a desirable regret upper bound for the \texttt{Fed-PNE} algorithm because the first term on the right-hand side only depends on $M$ and it is a cost due to federation across all the clients. When $T$ is sufficiently large compared with $M$ \footnote{Specifically, when $T^{d+1} > M^{1-({d+2})/({2\log_{k}\rho})}$ is satisfied.}, the second term dominates the bound and it depends sub-linearly on both the number of rounds $T$ and the number of agents $M$, which means that the algorithm converges to the optimum of the global objective. Moreover, the average cumulative regret of each client is of order $\widetilde{\mathcal{O}}\left(M^{-\frac{1}{d+2}}T^{\frac{d+1}{d+2}}\right)$, which represents that increasing the number of clients helps reducing the regret of each client, and thus validates the effectiveness of federation. Compared with the regret of centralized $\mathcal{X}$-armed bandit algorithms, i.e., $\widetilde{\mathcal{O}}\left(T^{\frac{d+1}{d+2}}\right)$  \citep{bubeck2011X, azar2014online}, the average regret bound of our algorithm is smaller when $M$ is large, which means that our algorithm is faster.
\end{remark}

\begin{remark}
When $T$ is relatively small, the first term in Theorem \ref{thm: regret_lower_bound} dominates the regret bound, yielding a super-linear dependence w.r.t. $M$ (but no dependence on $T$). Such a rate is mainly due to the lack of information and thus (potentially) inefficient sampling in the early stage, especially when there are too many clients. For example, when we explore the shallow layers, i.e., $h$ is small, in the partition at the beginning of the search, the total number of pulls of the node $\mathcal{P}_{h,i}$, i.e., $T_{h,i} =  M t_{m,h,i}$, could be much larger than the required threshold $\tau_h$.
\end{remark}

\begin{remark} 
\textbf{\upshape (Communication Rounds and Information)}
Moreover, the number of communication \textbf{\textit{rounds}} in Theorem \ref{thm: regret_upper_bound} only depends logarithmically on the time horizon $T$, showing that there are no frequent communications between the server and the clients during the federated learning process. Moreover, only the mean rewards are shared instead of all the rewards. Therefore, our algorithm successfully protects data confidentiality to certain extent and saves the communication cost. 
Similar dependence is observed in prior federated bandit works \citep{shi2021federateda}  \citep{huang2021federated}.

It's also worth mentioning that since the number of nodes $|\mathcal{K}^p|$ could increase when we increase the phase number $p$, a better measure of the communication cost is the amount of \textbf{\textit{information}} communicated instead of the number of \textbf{\textit{rounds}}. In this measure, the communication cost depends on the near-optimality dimension $d$ (Assumption \ref{assumption: near-optimality dimension}). If $d = 0$, it is easy to show that the communicated information is also of logarithmic order $\widetilde{\mathcal{O}}(M \log T)$. As mentioned by prior research, $d=0$ is the most commonly observed case for blackbox objectives \citep{bubeck2011X, Valko13Stochastic}. However, when $d > 0$, the communicated information could be as large as ${\mathcal{O}}(M T^{\frac{d}{d+2}})$ because both $|\mathcal{K}^p|$ and $\tau_{h^p}$ can exponentially increase when we increase $p$. In Appendix \ref{subsec: communication_cost}, we show that such dependence on $T$ is unfortunately unavoidable by any algorithm that has the same regret rate as \texttt{Fed-PNE}, and thus our cost is already optimal.
\end{remark}

\begin{remark} 
\textbf{\upshape (Privacy)}
The privacy guarantee in the main text refers to the limited communications between the server and the clients as in \citet{shi2021federateda, shi2021federatedb, huang2021federated}, instead of the quantitative privacy measures such as differential privacy (DP) \citep{dwork2010differential}.
However, since \texttt{Fed-PNE} only requires communications of the average rewards in very few rounds, it would be easy to guarantee differential privacy by adding Laplacian/Gaussian noise to the rewards in the \texttt{Fed-PNE} algorithm. In Appendix \ref{app: DP-Fed-PNE}, we 
prove our claim by presenting the differentially-private version of our algorithm (\texttt{DP-Fed-PNE}) and its analysis. 
\end{remark}

\subsection{Regret Lower Bound}

To show the tightness of the regret bound in Theorem \ref{thm: regret_upper_bound}, we provide the following lower bound. 

\begin{theorem}
\label{thm: regret_lower_bound}
There exists an instance of the federated $\mathcal{X}$-armed bandit problem {satisfying Assumptions 2.2 and 2.4} such that the expected cumulative regret of any multi-client algorithm is lower bounded as $\mathbb{E}[R(T)] = \Omega(M^{\frac{d+1}{d+2}} T^{\frac{d+1}{d+2}})$.
\end{theorem}

\begin{remark}
The proof of the above theorem is provided in Appendix \ref{subsec: regret_lower_bound}. Theorem \ref{thm: regret_lower_bound} essentially claims an  $\Omega(M^{\frac{d+1}{d+2}}T^{\frac{d+1}{d+2}})$ regret lower bound for the $M$-client, $T$-round federated $\mathcal{X}$-armed bandit problem, even if we allow instantaneous and unlimited number of communications between the clients and the server, i.e., the clients and the server can communicate in every round about the reward of any $x_{m,t}$ they choose. Therefore, the regret upper bound in Theorem \ref{thm: regret_upper_bound} is asymptotically unimprovable if we ignore the logarithmic term $\mathcal{O}(\log(MT)^{\frac{1}{d+2}})$.
\end{remark}

\section{Experiments}
\label{sec: experiments}

\begin{figure*}
\centering
\hspace*{-1.5em}
\subfigure[\footnotesize {Garland}]{
  \centering
  \includegraphics[width=0.27\linewidth]{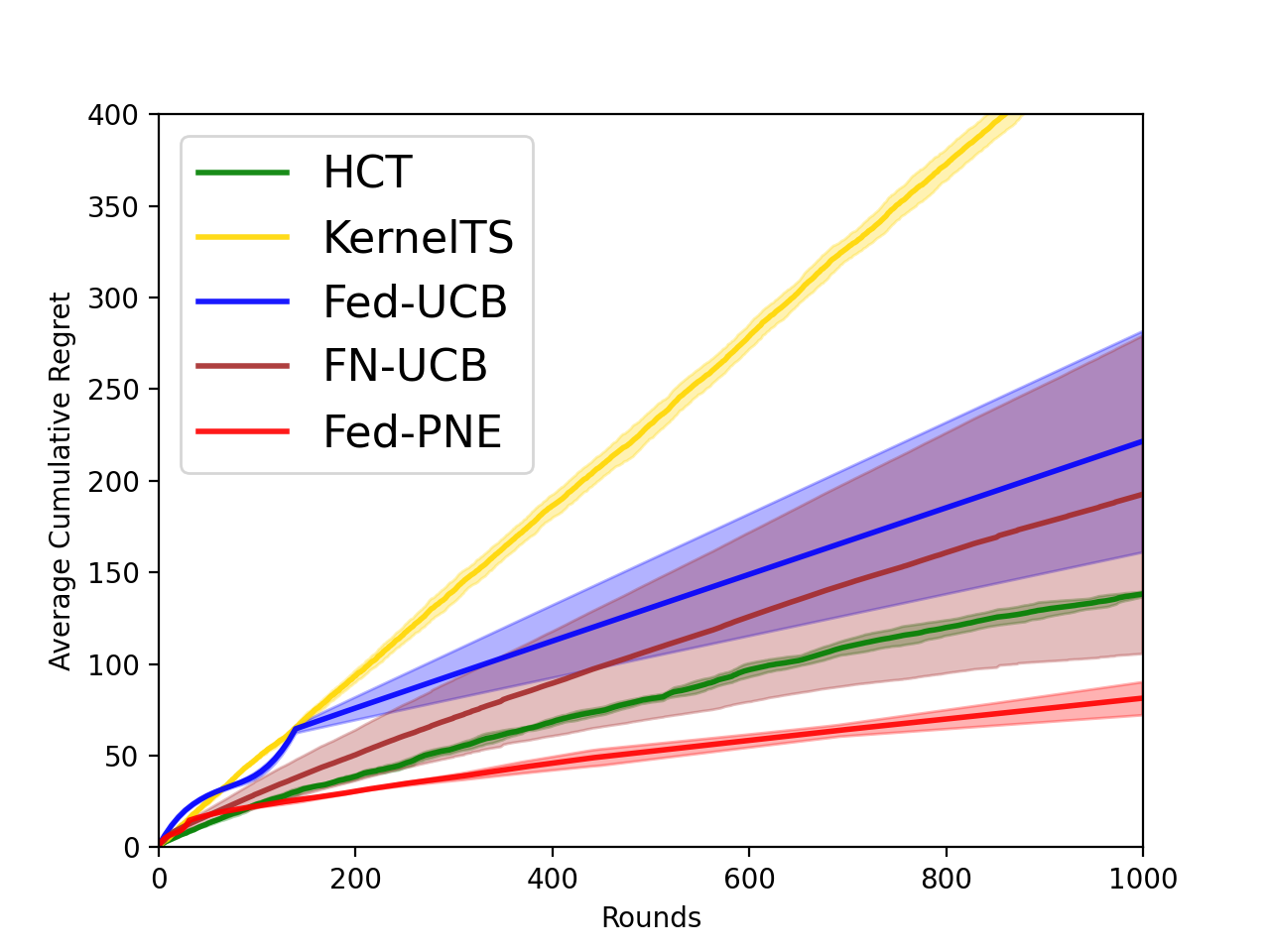}
  \label{fig: Garland}
}\hspace*{-1.5em}%
\subfigure[\footnotesize {DoubleSine}]{
  \centering
  \includegraphics[width=0.27\linewidth]{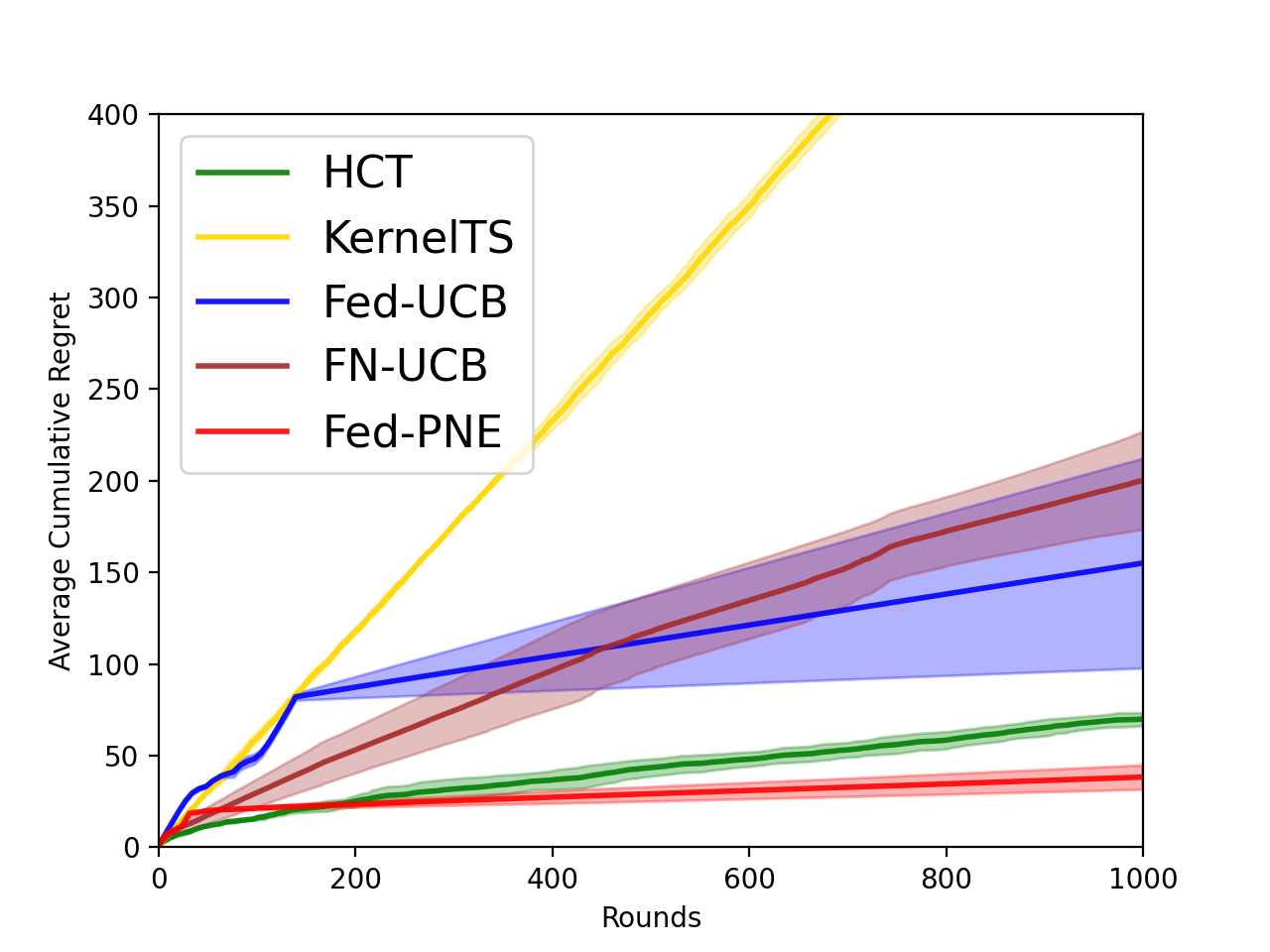}
    \label{fig: DoubleSine}
}\hspace*{-1.5em}%
\subfigure[\footnotesize Landmine]{
  \centering
  \includegraphics[width=0.27\linewidth]{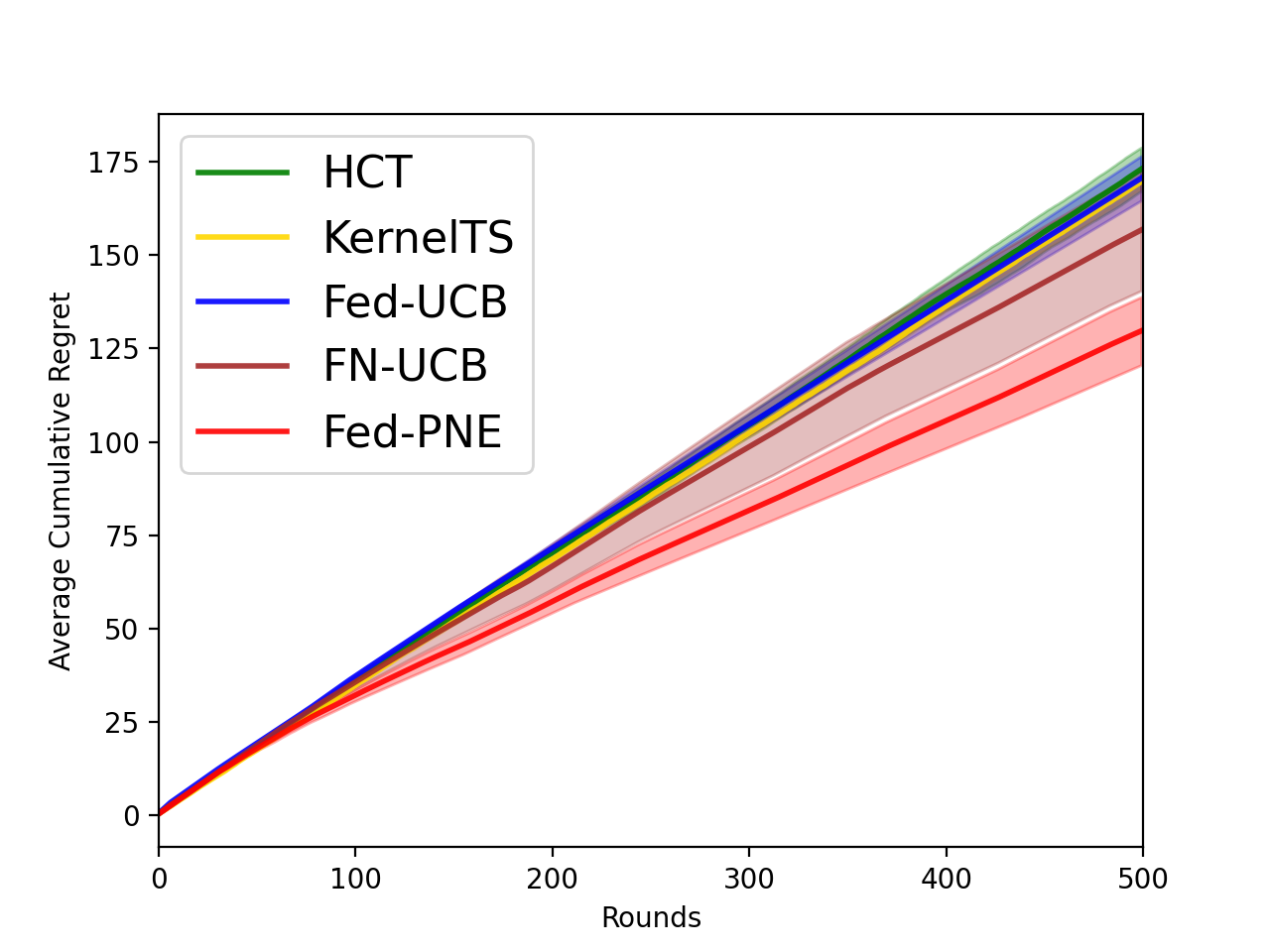}
  \label{fig: Landmine}
}\hspace*{-1.5em}%
\subfigure[\footnotesize COVID]{
  \centering
  \includegraphics[width=0.27\linewidth]{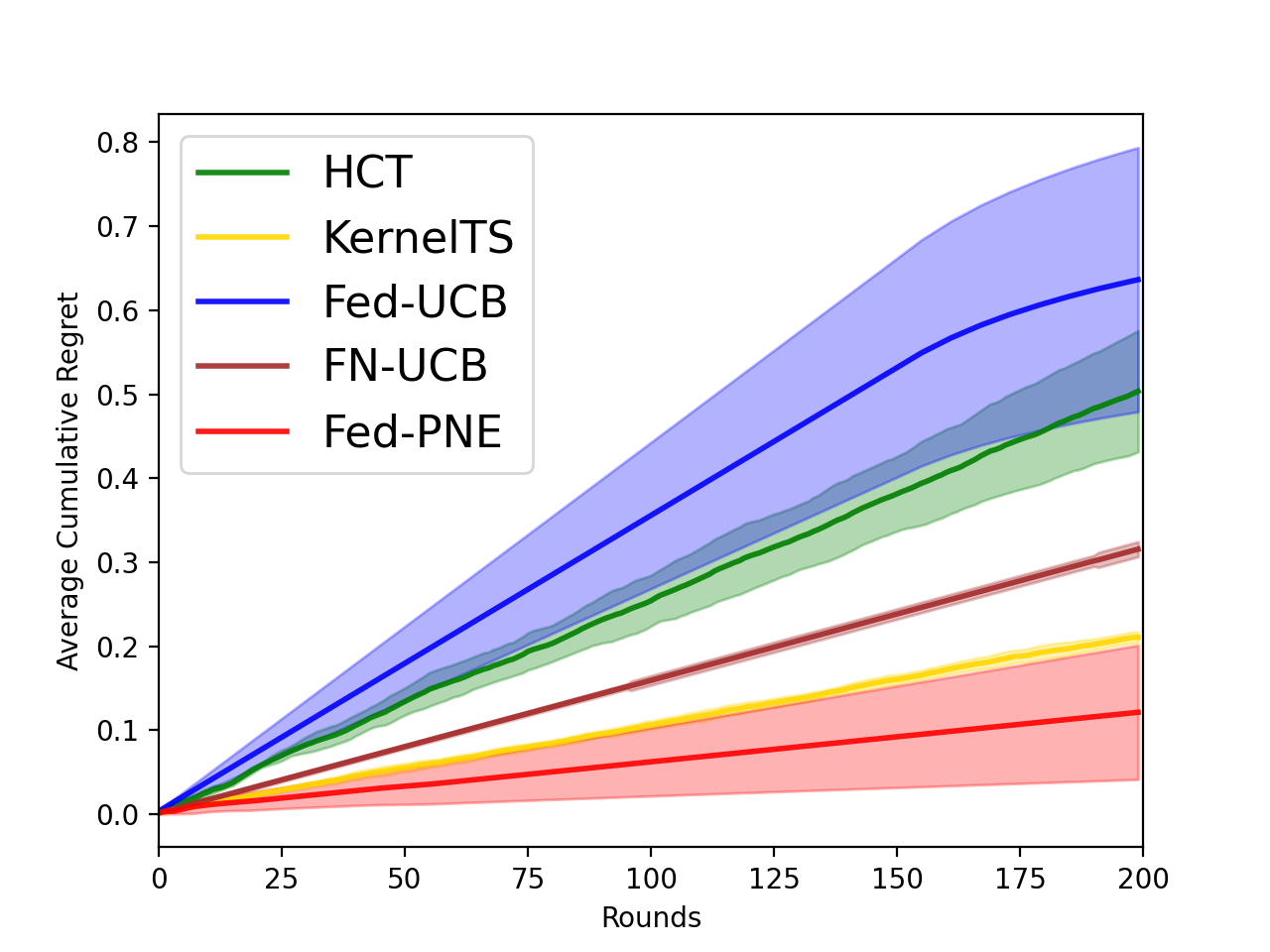}
  \label{fig: COVID}
}
\caption{Cumulative regret of different algorithms on the synthetic functions and the real-life datasets. Unlimited communications are allowed for centralized algorithms.
}
\vspace{-10pt}
\label{fig: experiments}
\end{figure*}

We empirically  evaluate the proposed \texttt{Fed-PNE} algorithm on both synthetic functions and real-world datasets. We compare \texttt{Fed-PNE} with centralized $\mathcal{X}$-armed bandit algorithm $\texttt{HCT}$ \citep{azar2014online}, centralized kernelized bandit algorithm \texttt{KernelTS} \citep{chowdhury2017kernelized}, federated multi-armed bandit algorithm \texttt{Fed1-UCB} \citep{shi2021federateda}, and federated neural bandit algorithm \texttt{FN-UCB} \citep{dai2023federated}. Additional details of algorithm implementations and more comparisons against other blackbox optimization algorithms such as Bayesian Optimization and Batched Bayesian Optimization algorithms,  are provided in Appendix \ref{app: experiment}.

\begin{remark}
For the federated algorithms (\texttt{Fed-PNE, Fed1-UCB, FN-UCB}), we plot the average cumulative regret per client against the rounds. For the centralized algorithms (\texttt{HCT}, \texttt{KernelTS}), we plot the cumulative regret on the global objective of each task against the number of evaluations. Such a comparison is fair in terms of overall computation resource, since the global objective itself is not directly accessible, and we can view one evaluation of global objective as the result of instant public communications of all local objective evaluations in one round. For all the curves presented in this section (and the numerical results in the appendix), they are averaged over 10 independent runs with shaded area standing for the 1 standard deviation. 
\end{remark}

\textbf{Synthetic Dataset.} We evaluate the algorithms on two synthetic functions that are commonly used in $\mathcal{X}$-armed bandit problem, which are the Garland function and the DoubleSine function, both defined on $\mathcal{X} = [0, 1]$. These two functions are well-known for their large number of local optimums. The randomly perturbed versions of these two functions are used as the local objective while the {averages of the local objectives} are used as the global objective. The average cumulative regret of different algorithms are provided in Figure \ref{fig: Garland} and \ref{fig: DoubleSine}. As can be observed in the figures, 
\texttt{Fed-PNE} has the smallest cumulative regret.

\textbf{Landmine Detection.} We federatedly tune the hyper-parameters of machine learning models fitted on the Landmine dataset \citep{liu2007semi}, where the features of different locations on multiple landmine fields extracted from radar images are used to detect the landmines. Following the setting of \citet{dai2020federated}, each client only has the access to the data of one random field, and trains a support vector machine with the RBF kernel parameter chosen from [0.01, 10] and the $L_2$ regularization parameter chosen from $[10^{-4}, 10]$. The local objectives and the global objective are the AUC-ROC scores on the local landmine field and all the landmine fields respectively.  The average cumulative regret of different algorithms are provided in Figure \ref{fig: Landmine}. As can be observed in the figures, our algorithm achieved smallest cumulative regret and thus the best performance.

\textbf{COVID-19 Vaccine Dosage Optimization.} In combat to the pandemic, we optimize the vaccine dosage in epidemiological models of COVID-19 to find the best fractional dosage for the overall population following \citet{Wiecek2022Testing}. Using fractional dosage of the vaccines will make them less effective, but at the same time more people get the chance of vaccination and thus can possibly accelerate the process of herd immunity. In our experimental setting, the local objectives are the final infectious rate of different countries/regions. Different countries have different parameters such as population size and the number of ICU units, and thus make the objectives heterogeneous. The results are shown in Figure \ref{fig: COVID}. Our algorithm also achieves the fastest convergence.

\section{Discussions and Conclusions}
\label{sec: conclusions}

In this work, we establish the framework of federated $\mathcal{X}$-armed bandit problem and propose the first algorithm for such problems. The proposed \texttt{Fed-PNE} algorithm utilizes the intrinsic structure of the global objective inside the hierarchical partitioning and achieves desirable regret bounds in terms of both the number of clients and the evaluation budget. Meanwhile it requires only logarithmic communications between the server and the clients, protecting the privacy of the clients. Both theoretical analysis and the experimental results show the advantage of \texttt{Fed-PNE} over centralized algorithms and prior federated multi-armed bandit algorithms. Many interesting future directions can be explored based on the framework proposed in this work. For example, other summary statistics of the client-wise data can potentially accelerate the proposed algorithm, such as the usage of empirical variance in \citet{li2021optimumstatistical}. Moreover, the current algorithm still needs a the weak lipschitzness assumption. Whether the weakest assumption in the literature of $\mathcal{X}$-armed bandit, i.e., the \textit{local smooth without a metric} assumption proposed by \citet{Grill2015Blackbox} can be used to prove similar regret guarantees remains a challenging open problem. 

\clearpage

\bibliographystyle{plainnat}
\bibliography{my_ref}

\begin{thebibliography}{35}
\providecommand{\natexlab}[1]{#1}
\providecommand{\url}[1]{\texttt{#1}}
\expandafter\ifx\csname urlstyle\endcsname\relax
  \providecommand{\doi}[1]{doi: #1}\else
  \providecommand{\doi}{doi: \begingroup \urlstyle{rm}\Url}\fi

\bibitem[Azar et~al.(2014)Azar, Lazaric, and Brunskill]{azar2014online}
Mohammad~Gheshlaghi Azar, Alessandro Lazaric, and Emma Brunskill.
\newblock Online stochastic optimization under correlated bandit feedback.
\newblock In \emph{International Conference on Machine Learning}, pages
  1557--1565. PMLR, 2014.

\bibitem[Bartlett et~al.(2019)Bartlett, Gabillon, and
  Valko]{bartlett2019simple}
Peter~L. Bartlett, Victor Gabillon, and Michal Valko.
\newblock A simple parameter-free and adaptive approach to optimization under a
  minimal local smoothness assumption.
\newblock In \emph{30th International Conference on Algorithmic Learning
  Theory}, 2019.

\bibitem[Bastani and Bayati(2020)]{bastani2020online}
Hamsa Bastani and Mohsen Bayati.
\newblock Online decision making with high-dimensional covariates.
\newblock \emph{Operations Research}, 68\penalty0 (1):\penalty0 276--294, 2020.

\bibitem[Bubeck et~al.(2011)Bubeck, Munos, Stoltz, and
  Szepesv{{\'a}}ri]{bubeck2011X}
S{{\'e}}bastien Bubeck, R{{\'e}}mi Munos, Gilles Stoltz, and Csaba
  Szepesv{{\'a}}ri.
\newblock $\chi$-armed bandits.
\newblock \emph{Journal of Machine Learning Research}, 12\penalty0
  (46):\penalty0 1655--1695, 2011.

\bibitem[Chen and Gallego(2022)]{chen2022APrimal}
Ningyuan Chen and Guillermo Gallego.
\newblock A primal–dual learning algorithm for personalized dynamic pricing
  with an inventory constraint.
\newblock \emph{Mathematics of Operations Research}, 0\penalty0 (0):\penalty0
  null, 2022.
\newblock \doi{10.1287/moor.2021.1220}.

\bibitem[Chowdhury and Gopalan(2017)]{chowdhury2017kernelized}
Sayak~Ray Chowdhury and Aditya Gopalan.
\newblock On kernelized multi-armed bandits.
\newblock In \emph{Proceedings of the 34th International Conference on Machine
  Learning}, volume~70 of \emph{Proceedings of Machine Learning Research},
  pages 844--853. PMLR, 06--11 Aug 2017.

\bibitem[Dai et~al.(2020)Dai, Low, and Jaillet]{dai2020federated}
Zhongxiang Dai, Bryan Kian~Hsiang Low, and Patrick Jaillet.
\newblock Federated bayesian optimization via thompson sampling.
\newblock In \emph{Advances in Neural Information Processing Systems},
  volume~33, pages 9687--9699. Curran Associates, Inc., 2020.

\bibitem[Dai et~al.(2021)Dai, Low, and Jaillet]{dai2021differentially}
Zhongxiang Dai, Bryan Kian~Hsiang Low, and Patrick Jaillet.
\newblock Differentially private federated bayesian optimization with
  distributed exploration.
\newblock In \emph{Advances in Neural Information Processing Systems},
  volume~34, pages 9125--9139. Curran Associates, Inc., 2021.

\bibitem[Dai et~al.(2023)Dai, Shu, Verma, Fan, Low, and
  Jaillet]{dai2023federated}
Zhongxiang Dai, Yao Shu, Arun Verma, Flint~Xiaofeng Fan, Bryan Kian~Hsiang Low,
  and Patrick Jaillet.
\newblock Federated neural bandits.
\newblock In \emph{The Eleventh International Conference on Learning
  Representations}, 2023.
\newblock URL \url{https://openreview.net/forum?id=38m4h8HcNRL}.

\bibitem[Dubey and Pentland(2020)]{dubey2020differentially}
Abhimanyu Dubey and Alex~Sandy Pentland.
\newblock Differentially-private federated linear bandits.
\newblock In \emph{Advances in Neural Information Processing Systems},
  volume~33, pages 6003--6014. Curran Associates, Inc., 2020.

\bibitem[Dwork et~al.(2010)Dwork, Naor, Pitassi, and
  Rothblum]{dwork2010differential}
Cynthia Dwork, Moni Naor, Toniann Pitassi, and Guy Rothblum.
\newblock Differential privacy under continual observation.
\newblock In \emph{STOC '10: Proceedings of the 42nd ACM symposium on Theory of
  computing}, pages 715--724. ACM, June 2010.

\bibitem[Feng et~al.(2021)Feng, Huang, and Wang]{Feng2021Lipschitz}
Yasong Feng, Zengfeng Huang, and Tianyu Wang.
\newblock Lipschitz bandits with batched feedback, 2021.

\bibitem[Frazier(2018)]{frazier2018tutorial}
Peter~I. Frazier.
\newblock A tutorial on bayesian optimization, 2018.
\newblock URL \url{https://arxiv.org/abs/1807.02811}.

\bibitem[Grill et~al.(2015)Grill, Valko, Munos, and Munos]{Grill2015Blackbox}
Jean-Bastien Grill, Michal Valko, Remi Munos, and Remi Munos.
\newblock Black-box optimization of noisy functions with unknown smoothness.
\newblock In \emph{Advances in Neural Information Processing Systems}. Curran
  Associates, Inc., 2015.

\bibitem[Huang et~al.(2021)Huang, Wu, Yang, and Shen]{huang2021federated}
Ruiquan Huang, Weiqiang Wu, Jing Yang, and Cong Shen.
\newblock Federated linear contextual bandits.
\newblock In \emph{Advances in Neural Information Processing Systems}, 2021.

\bibitem[Jones et~al.({1993})Jones, Perttunen, and
  Stuckman]{Jones1993Lipschitzian}
David Jones, Cary Perttunen, and Bruce Stuckman.
\newblock {Lipschitzian optimization without the Lipschitz constant}.
\newblock \emph{{Journal of Optimization Theory and Applications}},
  {79}\penalty0 ({1}):\penalty0 {157--181}, {Oct} {1993}.

\bibitem[Khodak et~al.(2021)Khodak, Tu, Li, Li, Balcan, Smith, and
  Talwalkar]{Khodak2021Federated}
Mikhail Khodak, Renbo Tu, Tian Li, Liam Li, Maria-Florina~F Balcan, Virginia
  Smith, and Ameet Talwalkar.
\newblock Federated hyperparameter tuning: Challenges, baselines, and
  connections to weight-sharing.
\newblock In \emph{Advances in Neural Information Processing Systems},
  volume~34, pages 19184--19197. Curran Associates, Inc., 2021.

\bibitem[Kleinberg et~al.(2008)Kleinberg, Slivkins, and
  Upfal]{kleinberg2008multi-armed}
Robert Kleinberg, Aleksandrs Slivkins, and Eli Upfal.
\newblock Multi-armed bandits in metric spaces.
\newblock In \emph{Proceedings of the Fortieth Annual ACM Symposium on Theory
  of Computing}, STOC '08, page 681–690, New York, NY, USA, 2008. Association
  for Computing Machinery.
\newblock ISBN 9781605580470.
\newblock \doi{10.1145/1374376.1374475}.

\bibitem[Lattimore and Szepesv{\'a}ri(2020)]{lattimore2020bandit}
Tor Lattimore and Csaba Szepesv{\'a}ri.
\newblock \emph{Bandit algorithms}.
\newblock Cambridge University Press, 2020.

\bibitem[Li et~al.(2020)Li, Song, and Fragouli]{li2020federated}
Tan Li, Linqi Song, and Christina Fragouli.
\newblock Federated recommendation system via differential privacy.
\newblock In \emph{2020 IEEE International Symposium on Information Theory
  (ISIT)}, pages 2592--2597. IEEE, 2020.

\bibitem[Li et~al.(2021)Li, Wang, Song, and Cheng]{li2021optimumstatistical}
Wenjie Li, Chi-Hua Wang, Qifan Song, and Guang Cheng.
\newblock Optimum-statistical collaboration towards general and efficient
  black-box optimization, 2021.

\bibitem[Li et~al.(2023)Li, Li, Honorio, and Song]{Li2023PyXAB}
Wenjie Li, Haoze Li, Jean Honorio, and Qifan Song.
\newblock Pyxab -- a python library for $\mathcal{X}$-armed bandit and online
  blackbox optimization algorithms, 2023.
\newblock URL \url{https://arxiv.org/abs/2303.04030}.

\bibitem[Liu et~al.(2007)Liu, Liao, and Carin]{liu2007semi}
Qiuhua Liu, Xuejun Liao, and Lawrence Carin.
\newblock Semi-supervised multitask learning.
\newblock In \emph{Advances in Neural Information Processing Systems},
  volume~20. Curran Associates, Inc., 2007.

\bibitem[McMahan et~al.(2017)McMahan, Moore, Ramage, Hampson, and
  y~Arcas]{mcmahan2017communication}
Brendan McMahan, Eider Moore, Daniel Ramage, Seth Hampson, and Blaise~Aguera
  y~Arcas.
\newblock Communication-efficient learning of deep networks from decentralized
  data.
\newblock In \emph{Artificial intelligence and statistics}, pages 1273--1282.
  PMLR, 2017.

\bibitem[Munos(2011)]{Munos2011Optimistic}
R\'{e}mi Munos.
\newblock Optimistic optimization of a deterministic function without the
  knowledge of its smoothness.
\newblock In \emph{Advances in Neural Information Processing Systems},
  volume~24. Curran Associates, Inc., 2011.

\bibitem[Shang et~al.(2019)Shang, Kaufmann, and Valko]{shang2019general}
Xuedong Shang, Emilie Kaufmann, and Michal Valko.
\newblock General parallel optimization a without metric.
\newblock In \emph{Algorithmic Learning Theory}, pages 762--788, 2019.

\bibitem[Shariff and Sheffet(2018)]{shariff2018Differentially}
Roshan Shariff and Or~Sheffet.
\newblock Differentially private contextual linear bandits.
\newblock In \emph{Advances in Neural Information Processing Systems},
  volume~31. Curran Associates, Inc., 2018.

\bibitem[Shi and Shen(2021a)]{shi2021federateda}
Chengshuai Shi and Cong Shen.
\newblock Federated multi-armed bandits.
\newblock \emph{Proceedings of the AAAI Conference on Artificial Intelligence},
  35\penalty0 (11):\penalty0 9603--9611, May 2021a.

\bibitem[Shi et~al.(2021b)Shi, Shen, and Yang]{shi2021federatedb}
Chengshuai Shi, Cong Shen, and Jing Yang.
\newblock Federated multi-armed bandits with personalization.
\newblock In \emph{Proceedings of The 24th International Conference on
  Artificial Intelligence and Statistics}, volume 130 of \emph{Proceedings of
  Machine Learning Research}, pages 2917--2925. PMLR, 13--15 Apr 2021b.

\bibitem[Tossou and Dimitrakakis(2016)]{Tossou2016Algorithms}
Aristide C.~Y. Tossou and Christos Dimitrakakis.
\newblock Algorithms for differentially private multi-armed bandits.
\newblock In \emph{Proceedings of the Thirtieth AAAI Conference on Artificial
  Intelligence}, AAAI'16, page 2087–2093. AAAI Press, 2016.

\bibitem[Valko et~al.(2013)Valko, Carpentier, and Munos]{Valko13Stochastic}
Michal Valko, Alexandra Carpentier, and Rémi Munos.
\newblock Stochastic simultaneous optimistic optimization.
\newblock In \emph{Proceedings of the 30th International Conference on Machine
  Learning}, volume~28 of \emph{Proceedings of Machine Learning Research},
  pages 19--27. PMLR, 17--19 Jun 2013.

\bibitem[Wang et~al.(2022)Wang, Li, Cheng, and Lin]{Wang2022Federated}
Chi-Hua Wang, Wenjie Li, Guang Cheng, and Guang Lin.
\newblock Federated online sparse decision making, 2022.

\bibitem[Wang et~al.(2018)Wang, Gehring, Kohli, and Jegelka]{wang2018batched}
Zi~Wang, Clement Gehring, Pushmeet Kohli, and Stefanie Jegelka.
\newblock Batched large-scale bayesian optimization in high-dimensional spaces.
\newblock In \emph{Proceedings of the Twenty-First International Conference on
  Artificial Intelligence and Statistics}, pages 745--754. PMLR, 2018.

\bibitem[Wiecek et~al.(2022)Wiecek, Ahuja, Chaudhuri, Kremer, Gomes, Snyder,
  Tabarrok, and Tan]{Wiecek2022Testing}
Witold Wiecek, Amrita Ahuja, Esha Chaudhuri, Michael Kremer, Alexandre~Simoes
  Gomes, Christopher~M. Snyder, Alex Tabarrok, and Brandon~Joel Tan.
\newblock Testing fractional doses of covid-19 vaccines.
\newblock \emph{Proceedings of the National Academy of Sciences}, 119\penalty0
  (8):\penalty0 e2116932119, 2022.

\bibitem[Zhu et~al.(2021)Zhu, Zhu, Liu, and Liu]{zhu2021federated}
Zhaowei Zhu, Jingxuan Zhu, Ji~Liu, and Yang Liu.
\newblock Federated bandit: A gossiping approach.
\newblock In \emph{Abstract Proceedings of the 2021 ACM
  SIGMETRICS/International Conference on Measurement and Modeling of Computer
  Systems}, pages 3--4, 2021.

\end{thebibliography}

\clearpage
\onecolumn
\appendix

\centerline{\textbf{\LARGE Appendix to ``Federated $\mathcal{X}$-Armed Bandit" }}

\section{Additional Related Works.}
\label{sec: related_works}

\textbf{Federated bandits.} 
Most recent works of federated bandits focus on the case where the number of arms is finite or the reward function is linear. For example, \citet{shi2021federateda} and \cite{shi2021federatedb} propose the \texttt{Fed-UCB}-type algorithms for the multi-armed bandit (with personalization) problem to construct efficient client-server communications. \citet{zhu2021federated, li2020federated} propose to make use of differential privacy to protect the user information of each client in federated bandit. For federated linear contextual bandits, different algorithms such as \citet{dubey2020differentially, huang2021federated} utilize distinct ways to reconstruct the global contextual parameter and achieve sublinear regret with little communication cost. Very recently, \citet{Wang2022Federated} extended the LASSO bandit algorithm \citep{bastani2020online} to federated high dimensional bandits. As far as we are aware of, our work is the first progress to discuss federated $\mathcal{X}$-armed bandit or continuum-armed bandit.

\textbf{$\mathcal{X}$-armed bandits.} Since the creation of the \texttt{Zooming} algorithm \citep{kleinberg2008multi-armed}, $\mathcal{X}$-armed bandit has become a heated line of research. Algorithms such as \texttt{HOO, HCT, VHCT} provide cumulative regret bounds for the stochastic reward feedback setting \citep{bubeck2011X, azar2014online, li2021optimumstatistical}. Apart from these works, some algorithms also discuss the case where the reward has no noise, such as \texttt{DiRect} \citep{Jones1993Lipschitzian}, \texttt{DOO} \citep{Munos2011Optimistic}, \texttt{SOO}\citep{Munos2011Optimistic}, and \texttt{SequOOL} \citep{bartlett2019simple}.  Another set of research works focuses on minimizing the \textit{simple regret} of the optimization algorithms at the final round, and proposes algorithms such as \texttt{POO} \citep{Grill2015Blackbox},  \texttt{GPO} \citep{shang2019general}, and \texttt{StroquOOL} \citep{bartlett2019simple}.  The theoretical results of these algorithms are not directly comparable to our work since we focus on cumulative regret.

\textbf{Federated hyper-parameter optimization.} Recently, a few works have analyzed the problem of federated hyper-parameter optimization using different approaches. \citet{dai2020federated, dai2021differentially} have proposed some modified versions of Thompson sampling and analyzed the problem using Bayesian Optimization algorithms. However, these algorithms require strong assumptions and the regret bounds of these algorithms do not show clear dependence on the number of clients \citep{dai2020federated, dai2021differentially}.  \citet{Khodak2021Federated} have proposed \texttt{FedEx} for optimizing hyper-parameters of algorithms such as \texttt{Fed-Avg} and analyzed its regret in convex problems. Compared with these works, our work is much more general in three senses: (1) We require very weak assumptions on the objectives for the \texttt{Fed-PNE} algorithm to work; (2) Our work can be applied to non-hyper-parameter tuning problems such as medicine dosage prediction, which do not involve algorithms such as \texttt{Fed-Avg}; (3) We provide the first regret bound with clear dependence on $M$ under the weak assumptions.

\newenvironment{list2}{
  \begin{list}{$\bullet$}{  
      \setlength{\itemsep}{5pt}
      \setlength{\parsep}{5pt} \setlength{\parskip}{0in}
      \setlength{\topsep}{5pt} \setlength{\partopsep}{0in}
      \setlength{\leftmargin}{30pt}}}{\end{list}}

\section{Notations and Useful Lemmas}
\label{app: notaions_and_lemmas}

\subsection{Notations}

Here we list all the notations used in the proof of our cumulative regret bound:
\begin{list2}
    \item $P_t$ : The set of phases $\{1, 2, \cdots p, \cdots \}$ that are completed up to time $t$.
    \item $\mathcal{T}^p$: The time steps of phase $p$.
    \item $\mathcal{K}^p$: The set of un-eliminated/pre-eliminated nodes in phase $p$.
    \item $\mathcal{E}^p$: The set of nodes to be eliminated at the end of phase $p$.
    \item $\overline{\mathcal{K}}^p$: The set of post-eliminated nodes in phase $p$, i.e., $\overline{\mathcal{K}}^p = \mathcal{K}^p \setminus \mathcal{E}^p$.
    \item $(h^p, i^p)$: the depth and index of the node $\mathcal{P}_{h^p, i^p}$ chosen by the server at phase $p$ from $\mathcal{K}^p$.
    \item $(h^*, i^*)$:  the depth and index of the node $\mathcal{P}_{h^*, i^*}$ that contains the (one of the) global maximizer $x^*$ on depth $h^*$.
    \item $t_{m,h,i}$: the number of times the node $\mathcal{P}_{h,i}$ is sampled from client $m$, which is chosen to be $t_{m,h,i} = \left \lceil  \frac{\tau_{h}}{M} \right \rceil$ in the \texttt{FedPNE} algorithm.
    \item $\tau_h$: the minimum required number of samples needed for a node on depth $h$, defined below.
    \item $T_{h,i}$: the number of times the node $\mathcal{P}_{h,i}$ is sampled, which is chosen to be $T_{h,i} =  \left \lceil  \frac{\tau_{h}}{M} \right \rceil M $ in the \texttt{FedPNE} algorithm.
\end{list2}

\textbf{The threshold for every depth.} 
The number of times $\tau_{h}$ needed for the statistical error (the UCB term) of every node on depth $h$ to be better than the optimization error is the solution to
  
\begin{equation}
    \nu_1 \rho^h \approx c \sqrt{\frac{\log(c_1 T/\delta)}{\tau_h}},
\end{equation}
  
which is equivalent as the following choice of the threshold 
\begin{equation}
\label{eqn: tau_upper_lower_bound}
  \frac{c^2 }{  \nu_1^2}  \rho^{-2h}  \leq \tau_h = \left\lceil \frac{c^2 \log(c_1T/\delta)}{  \nu_1^2}  \rho^{-2h} \right \rceil \leq 2  \frac{c^2 \log(c_1T/\delta)}{  \nu_1^2}  \rho^{-2h}.
\end{equation}
  
Notably, this choice of the threshold is the same as the threshold value in the \texttt{HCT} algorithm \citep{azar2014online}. In other words, we design our algorithm so that the samples are from different clients uniformly and thus the estimators are unbiased, and at the same time we minimize the unspent budget due to such distribution. There is still some (manageable) unspent budget due to the floor operation in the computation of $t_{m,h,i}$. However because of the expansion criterion (line 5-6) in \texttt{Fed-PNE}, we are able to travel to very deep layers inside the partition very fast when there are a lot of clients, and thus \texttt{Fed-PNE} is faster than single-client $\mathcal{X}$-armed bandit algorithms.

\subsection{Supporting Lemmas}

\begin{lemma} 
\label{lem: hoeffding}
\textbf{\upshape (Hoeffding's Inequality)}
    Let $X_{1}, \ldots, X_{n}$ be independent random variables such that $a_{i} \leq X_{i} \leq b_{i}$ almost surely. Consider the sum of these random variables, $S_{n}=X_{1}+\cdots+X_{n}$.
Then for all $t>0$, we have
  
\begin{equation}
\begin{aligned}
\nonumber
 \mathbb{P}\left(\left|S_{n}-\mathbb{E}\left[S_{n}\right]\right| \geq t\right) \leq 2 \exp \left(-\frac{2 t^{2}}{\sum_{i=1}^{n}\left(b_{i}-a_{i}\right)^{2}}\right).
\end{aligned}
\end{equation}
Here $\mathbb{E}\left[S_{n}\right]$ is the expected value of $S_{n}$.
\end{lemma}

\begin{lemma}
    \label{lem: good_event}
    \textbf{\upshape (High Probability Event)} Define the ``good" event $E_t$ as
\begin{equation}
    E_t = \left\{ \forall p \in P_t, \forall (h,i) \in \mathcal{K}^p, \forall T_{h,i} \in [MT], | f(x_{h,i}) - \widehat{\mu}_{h,i}| \leq  c \sqrt{\frac{\log(c_1 T/\delta)}{T_{h,i}}} \right\}
\end{equation}
 where the right hand side is exactly the confidence bound  $b_{h,i}$ for the node $\mathcal{P}_{h,i}$ and $c\geq 2, c_1 \geq (2M)^{1/8}$ are two constants. Then for any fixed round $t$, we have
$
     \mathbb{P}(E_t) \geq 1 - \delta/T^6
$
\end{lemma}
  
\textbf{Proof}.
For every $p \in P_t$ and every $(h,i) \in \mathcal{K}^p$, note that the node $\mathcal{P}_{h,i}$ is sampled the same number of times $(t_{m,h,i})$ independently from every local objective $m$, therefore by the Hoeffding's inequality (Lemma \ref{lem: hoeffding}), for any $x>0$, we have
\begin{equation}
  \mathbb{P}\left( \left| \sum_{m \in [M]} \sum_{t \in [t_{m,h,i}]} r_{m, h, i, t} -  \sum_{m \in [M]} t_{m,h,i} f_m (x_{h,i}) \right| \geq x \right) \leq 2 \exp \left(-\frac{2 x^{2}}{T_{h,i}}\right).
\end{equation}

If we take the average over the $M$ clients and the $t_{m,h,i}$ samples in the two summation terms inside the probability expression, we get that for any $x > 0$
\begin{equation}
  \mathbb{P}\bigg( \left | f(x_{h,i}) - \widehat{\mu}_{h,i} \right| \geq x \bigg) \leq 2 \exp \left(-{2T_{h,i} x^{2}}\right).   
\end{equation}
  
Therefore by the union bound, the probability of the event ${E}_t^c$ can be bounded as
\begin{equation}
\begin{aligned}
\mathbb{P} \left ({E}_{t}^{\mathrm{c}}\right)
&\leq \sum_{p \in P_t } \sum_{(h, i) \in \mathcal{K}^p} \sum_{T_{h,i}=1}^{MT} \mathbb{P}\bigg(| f(x_{h,i}) - \widehat{\mu}_{h,i}| >  b_{h,i} \bigg)\\
&\leq \sum_{p \in P_t } \sum_{(h, i) \in \mathcal{K}^p} 2MT \exp \bigg(- 2 T_{h,i} b_{h,i}^{2} \bigg)\\
& = 2MT \exp \bigg(-2c^2 \log(c_1 T/\delta) \bigg) \left ( \sum_{p \in P_t} |\mathcal{K}^p| \right) \\
&\leq 2 MT^2 \left(\frac{\delta}{c_1 T} \right)^{2c^2} \leq \frac{\delta}{T^6}.
\end{aligned}
\end{equation}
  
where the third inequality follows from the fact that the number of nodes is always smaller than $T$ since every client visits every node at least once.  \hfill $\square$

\begin{lemma}
    \label{lem: optimal_arm_not_eliminated}
    \textbf{(\upshape Optimal Node is Never Eliminated).} Under the high probability event ${E}_t$ at time $t \in \mathcal{T}^p$, the node that contains the global optimum $x^*$ of $f(x)$ at depth level $h_t$, denoted by $\mathcal{P}_{h^{*}_t, i^{*}_t}$, is never eliminated in Algorithm \ref{alg: server}, i.e., $(h^{*}_t, i^{*}_t) \notin \mathcal{E}^p$.
\end{lemma}
\textbf{Proof}. 
Under the high probability event ${E}_t$, we have the following inequality for every node $\mathcal{P}_{h, i} \in \mathcal{K}^p$
\begin{equation}
\begin{aligned}
      | f(x_{h,i}) - \widehat{\mu}_{h,i}| \leq b_{h,i} = c \sqrt{\frac{\log(c_1 T/\delta)}{T_{h,i}}}.
\end{aligned}
\end{equation}
  
Therefore we have the following set of inequalities at the end of each phase $p$ for $\mathcal{P}_{h^{*}_t, i^{*}_t}$.
\begin{equation}
\begin{aligned}
       \widehat{\mu}_{h_t^*,i_t^*} + \phi(h_t^*) + b_{h_t^*,i_t^*}  &\geq  f(x_{h_t^*,i_t^*}) +  \phi(h_t^*) \\
       & \geq f^* \geq f(x_{h^p, i^p}) \\
       &\geq \widehat{\mu}_{h^p, i^p} - b_{h^p, i^p},
\end{aligned}
\end{equation}
  
where the second inequality follows from the local smoothness property of the objective. Therefore we conclude that  $\mathcal{P}_{h^{*}_t, i^{*}_t} \notin \mathcal{E}^p$ and the optimal node is never eliminated. \hfill $\square$

\begin{lemma}
    \label{lem: remaining_nodes}
    \textbf{(\upshape Remaining Nodes Are Near-Optimal).} Under the high probability event ${E}_t$ at time $t \in \mathcal{T}^{p}$, the representative point $x_{h,i}$ of every un-eliminated node $\mathcal{P}_{h, i} $  in phase $p-1$, i.e., $(h,i) \in \overline{\mathcal{K}}^{p-1}$, is at least $6\nu_1 \rho^{h}$-optimal, that is
    \begin{equation}
        f^* - f(x_{h, i}) \leq 6\nu_1 \rho^{h}, \forall (h,i) \in \overline{\mathcal{K}}^{p-1}.
    \end{equation}
\end{lemma}
\textbf{Proof}. This Lemma shares a similar proof logic as Lemma \ref{lem: optimal_arm_not_eliminated}. Under the high probability event ${E}_t$, we have the following inequality for every node $\mathcal{P}_{h, i}$ such that $ (h,i) \in \overline{\mathcal{K}}^{p-1}$
\begin{equation}
\begin{aligned}
      | f(x_{h,i}) - \widehat{\mu}_{h,i}| \leq b_{h,i} = c \sqrt{\frac{\log(c_1 T/\delta)}{T_{h,i}}}.
\end{aligned}
\end{equation}
Therefore the following set of inequalities hold 
\begin{equation}
\begin{aligned}
       &f(x_{h, i}) + \nu_1 \rho^{h} + 2 b_{h, i} \geq  \widehat{\mu}_{h, i} + \nu_1 \rho^{h} + b_{h, i} \\ 
       &\geq  \widehat{\mu}_{h^{p-1}, i^{p-1}} - b_{h^{p-1}, i^{p-1}} \geq  \widehat{\mu}_{h^{*}, i^{*}} - b_{h^{p-1}, i^{p-1}}  \\
       &\geq  f(x_{h^{*}, i^{*}})  - b_{h^{*}, i^{*}}  - b_{h^{p-1}, i^{p-1}} \\
       &\geq  f^* - \nu_1 \rho^{h}  - b_{h^{*}, i^{*}}  - b_{h^{p-1}, i^{p-1}},
\end{aligned}
\end{equation}
where the second inequality holds because $\mathcal{P}_{h,i}$ is not eliminated. The third inequality holds because of the optimality of the node $\mathcal{P}_{h^{p-1}, i^{p-1}}$ in Algorithm \ref{alg: server}, and the last one follows from the weak lipchitzness assumption (Assumption \ref{assumption: local_smoothness}). In conclusion, we have the following upper bound on the regret
\begin{equation}
\begin{aligned}
        f^* - f(x_{h, i}) \leq  2\nu_1 \rho^{h}  + 2 b_{h, i}  +  b_{h^{*}, i^{*}}  + b_{h^{p-1}, i^{p-1}} \leq 6\nu_1 \rho^{h}
\end{aligned}
\end{equation}
where the last inequality holds because in the phase $p-1$, we sample each node enough number of times ($T_{h,i}$ larger than the threshold $\tau_h$) so that $b_{h, i} \leq \nu_1 \rho^{h}$ and thus $b_{h, i}, b_{h^{*}, i^{*}}, b_{h^{p-1}, i^{p-1}}$ are all smaller than $\nu_1 \rho^{h}$.   \hfill $\square$

\begin{lemma}
\label{lem: optimality_lipschitz}
\textbf{\upshape (Lemma 3 in \citet{bubeck2011X})}
    For a node $\mathcal{P}_{h,i}$, define $f^*_{h,i} = \sup_{x\in \mathcal{P}_{h,i}} f(x)$ to be the maximum of the function on that region. Suppose that $f^* - f^*_{h,i} \leq c \nu_1 \rho^h$ for some $c \geq 0$, then all $x$ in $\mathcal{P}_{h,i}$ are $\max\{2c, c+1\} \nu_1 \rho^h$-optimal.
\end{lemma}
\textbf{Proof.} The proof is provided here for completeness. For any real positive number $\delta > 0$, we denote by $x_{h,i}^*(\delta) \in \mathcal{P}_{h,i}$ such that 
\begin{equation}
       f(x_{h,i}^*(\delta) ) \geq f^*_{h,i}  - \delta = f^* - ( f^* - f^*_{h,i} ) - \delta.
\end{equation}
By the weak Lipschitz property, we have that for all $y \in \mathcal{P}_{h,i}$,
\begin{equation}
\begin{aligned}
     f^* - f(y) &\leq f^* -  f(x_{h,i}^*(\delta) )  + \max \left \{  f^* -  f(x_{h,i}^*(\delta) ), \ell(x_{h,i}^*(\delta), y)  \right \} \\
     &\leq ( f^* - f^*_{h,i} ) + \delta + \max \left \{ ( f^* - f^*_{h,i} ) + \delta, \text{diam}  (\mathcal{P}_{h,i}) \right\}.
\end{aligned}
\end{equation}
Letting $\delta \rightarrow 0$ and substituting the bounds on the suboptimality and on the diameter of $\mathcal{P}_{h,i}$ (Assumption \ref{assumption: local_smoothness}) concludes the proof. \hfill $\square$

\begin{lemma}
\label{lem: length_of_phase}\textbf{\upshape (Upper Bounds on Length of Phases)} Given the choice of the threshold $\tau_h$ and the design in Algorithm \ref{alg: server}, there exists a constant $h_M = \mathcal{O}(\log M)$ such that for any fixed phase $p$, denote $h^p$ and $h^{p-1}$ to be the depth sampled by the algorithm at phase $p$ and $p-1$ respectively, then the following inequalities hold almost surely
\begin{equation}
 |\mathcal{T}^p| \leq  
 \left \{
\begin{aligned}
  &4k\rho^{-2} \left(\frac{M\nu_1^2}{ c^2 } \right)^{-\frac{1}{2\log_k \rho}} & \text{ if }  h^{p-1} < h_M;\\
          & \frac{4C c^2 \log(c_1T/\delta) \rho^{-2} k }{M  \nu_1^2}  \rho^{ - (d'+2)h^{p-1}} &  \text{ otherwise. }   \\
\end{aligned}
\right .
\end{equation}
\end{lemma}
\textbf{Proof.} By the design in our algorithm, the length of each (dynamic) phase $p$ should be 
  
\begin{equation}
\begin{aligned}
  |\mathcal{T}^p| &=    |\mathcal{K}^p| \left \lceil \frac{\tau_{h^p}}{M}  \right  \rceil=  k^{h^p - h^{p-1}} \left|\overline{\mathcal{K}}^{p-1} \right| \left \lceil \frac{\tau_{h^p} }{M}  \right \rceil \leq  k^{h^p - h^{p-1}} C(\rho^{h^{p-1}})^{-{d'}} \left \lceil \frac{ \tau_{h^p}  }{M} \right \rceil \\
\end{aligned}
\end{equation}
  
where $k^{h^p - h^{p-1}}$ denotes the $h^p - h^{p-1}$-th power of $k$ and similar notations are used for the other powers. The second equality holds because nodes in $\mathcal{K}^p$ are descendants of the un-eliminated nodes in $\overline{\mathcal{K}}^{p-1} $, and the size of $\overline{\mathcal{K}}^{p-1}$ is bounded by $C(\rho^{h^{p-1}})^{-d'}$ by the definition of near-optimality dimension (Assumption \ref{assumption: near-optimality dimension}) and Lemma \ref{lem: remaining_nodes}. Define $h_M$ to be the depth where $\tau_{h} \geq M$, then 
\begin{equation}
    h_M \leq \frac{1}{2}\log_{\rho^{-1}} \left(\frac{M\nu_1^2 }{c^2} \right).
\end{equation}
Then when $h^{p-1} \geq h_M$, $h^p - h^{p-1} = 1$ because $\tau_{h^{p}}  \geq \tau_{h^{p-1}} \geq M$ and $|\mathcal{K}^p| \geq 1$, which means that $\tau_{h^p} / M > 1$ and thus the algorithm will only increase the depth by 1. Therefore $\left \lceil \frac{ \tau_{h^p}  }{M} \right \rceil \leq 2 \frac{ \tau_{h^p}  }{M}$, and by the upper bound on $\tau_{h^p}$ we have
\begin{equation}
\begin{aligned}
  |\mathcal{T}^p|   &\leq \frac{4k^{h^p - h^{p-1}} C c^2 \log(c_1T/\delta) }{M  \nu_1^2}  \rho^{-2h^p - d'h^{p-1}}       \\
   & = \frac{4C c^2 \log(c_1T/\delta) }{M  \nu_1^2} \left(\rho^{-2} k\right)^{h^p - h^{p-1}}   \rho^{ - (d'+2)h^{p-1}} \\
   &\leq  \frac{4C c^2 \log(c_1T/\delta) \rho^{-2} k }{M  \nu_1^2}  \rho^{ - (d'+2)(h^{p-1})}.
\end{aligned}
\end{equation}
 In the other case when $h^{p-1}< h_M$, we have $\tau_{h^{p-1}} \leq M$ and thus $\tau_{h^p} \leq  2\rho^{-2} M $ because this number is already larger than $M$. Now since $|\mathcal{K}^p| \leq k^{h_M + 1}$, we have the following bound on $|\mathcal{T}^p|$
\begin{equation}
\begin{aligned}
  |\mathcal{T}^p| &=  |\mathcal{K}^p| \left \lceil \frac{\tau_{h^p} }{M}  \right  \rceil \leq  k \left(\frac{M\nu_1^2}{ c^2 } \right)^{-\frac{1}{2\log_k \rho}}  \left \lceil {2\rho^{-2}}  \right  \rceil \leq 4k\rho^{-2} \left(\frac{M\nu_1^2}{ c^2 } \right)^{-\frac{1}{2\log_k \rho}},
\end{aligned}
\end{equation}
which finishes all the proof. \hfill $\square$

\section{Main Proofs}
\label{app: Main_Proof}

In this section, we provide the proofs of the main theorem (Theorem \ref{thm: regret_upper_bound}) in this paper.

\textbf{Proof.} Let $\mathbb{I}_{A}$ denote whether the event $A$ is true, i.e., $\mathbb{I}_{A} = 1$ if $A$ is true and 0 otherwise. We first decompose the regret into two terms
\begin{equation}
\begin{aligned}
    R(T) & =\sum_{m=1}^M \sum_{t=1}^T \left( f^* - f(x_{m, h_t, i_t})\right) \\
    &=\sum_{m=1}^M \sum_{t=1}^T \left( f^* - f(x_{m, h_t, i_t})\right)   \mathbb{I}_{E_t} + \sum_{m=1}^M \sum_{t=1}^T \left( f^* - f(x_{m, h_t, i_t})\right)   \mathbb{I}_{E_t^c} \\
    &= R(T)^{E}  + R(T)^{E^c}.
\end{aligned}
\end{equation}
 
For the second term, note that we can bound its expectation as follows
\begin{equation}
\begin{aligned}
\label{eqn: bound_on_Regret_Ec}
      \mathbb{E}\left[R(T)^{E^c}\right] &= \mathbb{E}\left[\sum_{m=1}^M \sum_{t=1}^T \left( f^* - f(x_{m, h_t, i_t})\right)  \mathbb{I}_{E_t^c}\right]  \leq \sum_{m=1}^M \sum_{t=1}^T \mathbb{P}\left(E_t^c\right) \\
      &\leq \sum_{m=1}^M \sum_{t=1}^T (\delta/T^6) = \frac{M\delta}{T^5}.
\end{aligned}
\end{equation}
 
where the second inequality  follows from Lemma \ref{lem: good_event}. Now we bound the first term $R(T)^{E}$ in the decomposition under the event $\mathcal{E}_t$. In every phase $p > 0$, for the non-eliminated nodes in the previous phase, i.e., $\mathcal{P}_{h,i}$ such that $(h, i) \in \overline{\mathcal{K}}^{p-1}$ (and thus $h = h^{p-1}$), by Lemma \ref{lem: remaining_nodes} we have
\begin{equation}
    f^* - f(x_{h,i}) \leq 6 \nu_1 \rho^{h^{p-1}}.
\end{equation}

By Lemma \ref{lem: optimality_lipschitz}, since the set ${\mathcal{K}}^p$ is created by expanding $\overline{\mathcal{K}}^{p-1}$, therefore for the representative point $x_{h', i'}$ of the node $\mathcal{P}_{h', i'}$ such that $(h', i') \in \mathcal{K}^p$, we have the following upper bound on the suboptimality gap
\begin{equation}
    f^* - f(x_{h', i'}) \leq 12 \nu_1 \rho^{h^{p-1}}.
\end{equation}
 
Let $H > 0$ be a constant depth to be chosen later. Since the clients can only sample from the set $\mathcal{K}^p$, we know that the term $R(T)^{E}$ can be written into the following form
\begin{equation}
\begin{aligned}
R(T)^E &= \sum_{m=1}^M \sum_{t=1}^T \left( f^* - f(x_{m, h_t, i_t})\right)   \mathbb{I}_{E_t} \leq \sum_{m=1}^M  \sum_{p=0} \sum_{t\in \mathcal{T}^p} 12 \nu_1 \rho^{h^{p-1}}  \\
& \leq \underbrace{\sum_{m=1}^M  \sum_{p: h^{p-1}\leq H} 12 \nu_1\rho^{h^{p-1}} |\mathcal{T}^p| }_{(a)} +  \underbrace{\sum_{m=1}^M  \sum_{p: h^{p-1} \geq H} \sum_{t\in \mathcal{T}^p}  12 \nu_1\rho^{h^{p-1}}} _{(b)}.
\end{aligned}
\end{equation}
 
We bound the two terms separately as follows, for term (a), we use Lemma \ref{lem: length_of_phase} and bound it as
\begin{equation}
\begin{aligned}
(a) &\leq \sum_{h=0}^{h_M} 48 M k \nu_1 \rho^{-2} \left(\frac{M\nu_1^2}{ c^2 } \right)^{-\frac{1}{2\log_k \rho}} \rho^{h} + \sum_{h=0}^H   \frac{48 k C c^2 \log(c_1T/\delta) \rho^{d} }{ \nu_1}  \rho^{- h - dh}  \\
&\leq \frac{48 M k \nu_1 }{\rho - \rho^3} \left(\frac{M\nu_1^2}{ c^2 } \right)^{-\frac{1}{2\log_k \rho}} +  \frac{48 k C c^2 \log(c_1T/\delta) \rho^{d} }{ \nu_1}   \sum_{h=0}^H \rho^{-(d+1)h} \\
&\leq \frac{48 M k \nu_1 }{\rho - \rho^3} \left(\frac{M\nu_1^2}{ c^2 } \right)^{-\frac{1}{2\log_k \rho}} +  \frac{48 k C c^2 \log(c_1T/\delta) \rho^{d - 1} }{ \nu_1 (\rho^{-({d}+1)} - 1)}  {\rho^{-({d}+1)H}},
\end{aligned}
\end{equation}
 
 where $h_M$ is the smallest depth such that $\tau_{h_M} \geq M$ defined in Lemma \ref{lem: length_of_phase}. For the nodes that have depth larger than $H$ in term $(b)$, we have the following bound
\begin{equation}
\begin{aligned}
(b) \leq 12 \nu_1\rho^H MT.
\end{aligned}
\end{equation}
 
Balancing the size of the dominating terms in the upper bounds of (a) and (b), we have
\begin{equation}
\begin{aligned}
\label{eqn: bound_on_Regret_E}
R (T)^{E} &\leq  \frac{48 k C c^2 \log(c_1T/\delta) \rho^{d - 1} }{ \nu_1 (\rho^{-({d}+1)} - 1)}  {\rho^{-({d}+1)H}} + 12 \nu_1 \rho^H MT +\frac{48 M k \nu_1 }{\rho - \rho^3} \left(\frac{M\nu_1^2}{ c^2 } \right)^{-\frac{1}{2\log_k \rho}} \\
&\leq  2 \left(\frac{48 k C c^2 \log(c_1T/\delta) \rho^{d - 1} (12\nu_1)^{d+1}}{ \nu_1 (\rho^{-({d}+1)} - 1)} \right)^{\frac{1}{d+2}} (MT)^{\frac{d+1}{d+2}} + \frac{48 M k \nu_1 }{\rho - \rho^3} \left(\frac{M\nu_1^2}{ c^2 } \right)^{-\frac{1}{2\log_k \rho}}.
\end{aligned}
\end{equation}
 
The last inequality holds because of the following choice of $H$
\begin{equation}
    \rho^H = \left ( \frac{4 k C c^2 \log(c_1T/\delta) \rho^{d - 1} }{MT \nu_1^2 (\rho^{-({d}+1)} - 1)} \right)^{\frac{1}{d+2}}.
\end{equation}

Combining Eqn. \ref{eqn: bound_on_Regret_Ec} with Eqn. \ref{eqn: bound_on_Regret_E}, we get the following upper bound on the expected cumulative regret of our algorithm by taking the confidence term $\delta = {1}/{M}$
\begin{equation}
\begin{aligned}
      \mathbb{E}[R (T)] &\leq 2 \left(\frac{48 k C c^2 \rho^{d - 1} (12\nu_1)^{d+1}}{ \nu_1 (\rho^{-({d}+1)} - 1)} \right)^{\frac{1}{d+2}} (MT)^{\frac{d+1}{d+2}} \log(c_1MT)^{\frac{1}{d+2}}+ \frac{1}{T^5} \\
      &\qquad + \frac{48 k \nu_1 }{\rho - \rho^3} \left(\frac{\nu_1^2}{ c^2 } \right)^{-\frac{1}{2\log_k \rho}} M^{1- \frac{1}{2\log_k \rho}}  \\
      & =  C_1 M^{1-\frac{1}{2\log_k \rho}} + C_2 \left( M^{\frac{d+1}{d+2}}  T^{\frac{d+1}{d+2}} (\log MT)^{\frac{1}{d+2}} \right),
\end{aligned}
\end{equation}
where in the first equality $C_1$ and $C_2$ are two absolute constants, specifically
\begin{equation}
    C_1 = \frac{48 k \nu_1 }{\rho - \rho^3} \left(\frac{\nu_1^2}{ c^2 } \right)^{-\frac{1}{2\log_k \rho}} + 1, C_2 = 2 \left(\frac{48 k C c^2 \rho^{d - 1} (12\nu_1)^{d+1}}{ \nu_1 (\rho^{-({d}+1)} - 1)} \right)^{\frac{1}{d+2}} \left(1+ \log c_1\right).
\end{equation}
where $c$ and $c_1$ are defined in \ref{lem: good_event}. Note that the first term in the regret bound does not depend on the total number of rounds $T$, and thus should be a negligible constant when $T$ is sufficiently large. Therefore our regret scales as $\mathcal{O} \left( M^{\frac{d+1}{d+2}}  T^{\frac{d+1}{d+2}} (\log MT)^{\frac{1}{d+2}} \right)$ asymptotically. \hfill $\square$
\subsection{Proof of Regret Lower Bound (Theorem \ref{thm: regret_lower_bound})}
\label{subsec: regret_lower_bound}

\textbf{Proof}. The construction of the regret lower bound can be achieved by considering the special case of our federated $\mathcal{X}$-armed bandit where all local objectives are the same as the global objective, and any rounds of communications between the $M$ clients are allowed. In that case, the problem becomes finding the optimal action for $M$ clients on one objective $f$ in each round. Note that it is equivalent to a ``batched $\mathcal{X}$-armed bandit" problem where one client optimizes the objective $f$, but can only receive the rewards of the chosen actions at a particular grid of time points \citep{Feng2021Lipschitz}, instead of at each time point. Particularly, our special case is equivalent to a uniform grid on $[0, MT]$ with interval width $M$ and batch size $T$. Now utilizing the lower bound in batched $\mathcal{X}$-armed bandit problem, we are able to provide a lower bound for the federated $\mathcal{X}$-armed bandit problem. 

Specifically, for any policies $[\pi_1, \pi_2, \cdots, \pi_{M}]$ employed by the $M$ clients in the federated $\mathcal{X}$-armed bandit problem with equal local objectives $f_m = f, \forall m \in [M]$, we can construct a batched $\mathcal{X}$-armed bandit problem on the same objective with a static grid $[0=t_0<t_1 < t_2 < \cdots < t_T = MT]$, where $t_i = Mi$. Now we define a policy $\Pi$ on the batched $\mathcal{X}$-armed bandit setting $\Pi(t) = \pi_{t \mod M}(\lceil t/M \rceil)$, then we know that

\begin{equation}
      \mathbb{E}_{[\pi_1, \pi_2, \cdots, \pi_{M}]}\left[R^{Fed}(T)\right] =    \mathbb{E}_{\Pi}\left[R^{Batch}(MT)\right] 
\end{equation}

By Lemma \ref{lem: batched_bandit_lower_bound} and the above argument, we can set $t_i = Mi$, $B=T$, and $n=MT$ in Lemma \ref{lem: batched_bandit_lower_bound}, then we know that
\begin{equation}
      \mathbb{E}_{[\pi_1, \pi_2, \cdots, \pi_{M}]}\left[R^{Fed}(T)\right] \geq \frac{1}{32 e^{\frac{1}{16}}}  \frac{MT}{[M(T-1)]^{\frac{1}{d+2}}} \geq  \frac{1}{32 e^{\frac{1}{16}}} M^{\frac{d+1}{d+2}} T^{\frac{d+1}{d+2}} = \Omega(M^{\frac{d+1}{d+2}} T^{\frac{d+1}{d+2}})
\end{equation}
Note that the above regret lower bound is expected since in the ``batched $\mathcal{X}$-armed bandit" case, the client receives less information compared with the original $\mathcal{X}$-armed bandit problem, and in the $\mathcal{X}$-armed bandit case, \citet{bubeck2011X} has proved a $\Omega(M^{\frac{d+1}{d+2}} T^{\frac{d+1}{d+2}})$ lower bound. Therefore, we would expect the lower bound in the batched $\mathcal{X}$-armed bandit problem to be not smaller than the bound in \citet{bubeck2011X}. \hfill $\square$

\begin{remark}
 A subtle difference between the upper bound in Theorem \ref{thm: regret_upper_bound} and the lower bound in Theorem \ref{thm: regret_lower_bound} is that $d$ represents the near optimality dimension (Assumption\ref{assumption: near-optimality dimension}) in the upper bound, but the dimension of the arm space in the lower bound. As mentioned in Theorem 12 in \citet{bubeck2011X} and the discussions in \citet{kleinberg2008multi-armed}, the upper bound can be easily converted to have the same notations as the lower bound without the need to change the proofs too much, and the two bounds match up to logarithmic terms. In that case, the near optimality dimension is the same as the dimension.
\end{remark}

\begin{lemma}
\label{lem: batched_bandit_lower_bound}
\textbf{\upshape (Equation before (26) in \citet{Feng2021Lipschitz})}
Given a static grid $[0=t_0<t_1 < t_2 < \cdots < t_B = n]$, there exists a batched Lipschitz bandit ($\mathcal{X}$-armed bandit) problem instance with $n$ total rounds such that for any policy, the expected cumulative regret satisfies
\begin{equation}
    \mathbb{E}\left[R(n)\right] \geq \frac{1}{32 e^{\frac{1}{16}}} \max \left\{t_1, \frac{t_2}{t_1^{\frac{1}{d+2}}}, \cdots, \frac{t_B}{t_{B-1}^{\frac{1}{d+2}}}\right\}
\end{equation}
where $d$ is the dimension of the arm space.
\end{lemma}

\subsection{Communication Rounds and Information}
\label{subsec: communication_cost}

\textbf{Proof.} In order to analyze the number of communication rounds between the server and the clients during the federated learning process, we need to bound the phase lengths from below instead of from above. Note that we have the lower bound on $\tau_{h^p}$ in Eqn. \ref{eqn: tau_upper_lower_bound} and $ |\mathcal{K}^p| \geq k$,  we have the following lower bound on $|\mathcal{T}^p|$
\begin{equation}
    |\mathcal{T}^p| =  |\mathcal{K}^p| \left \lceil \frac{\tau_{h^p}}{M}  \right  \rceil \geq \frac{k c^2 \rho^{-2h^p}  }{M \nu_1^2}.
\end{equation}
 
For simplicity, we consider a fixed number of phases. Let $P$ to be the total number of phases (and thus the rounds of communications), the total length of the algorithm would satisfy
\begin{equation}
    T = \sum_{p=1}^P |\mathcal{T}^p| \geq \sum_{p=1}^P \frac{k c^2 \rho^{-2h^p}  }{M \nu_1^2} \geq \frac{k c^2 \rho^{-2P}  }{M \nu_1^2},
\end{equation}
where the last inequality holds because $h_p \geq p, \forall p \geq 1$. Therefore, we get $P = \mathcal{O}(\log MT)$, specifically,
\begin{equation}
   P \leq \left( \log \left({\rho^{-2}} \right) \right)^{-1} \log \left(\frac{MT\nu_1^2}{kc^2} \right).
\end{equation}
 
Hence when $T$ is sufficiently large, the communication cost of \texttt{Fed-PNE} is of size 
\begin{equation}
    MP \leq M \left( \log \left({\rho^{-2}} \right) \right)^{-1}  \log \left(\frac{MT\nu_1^2}{kc^2} \right) = \widetilde{\mathcal{O}}(M \log T).
\end{equation}

Similar to the federated multi-armed bandit papers such as \citet{shi2021federateda, shi2021federatedb, huang2021federated}, we have measured the communication cost in terms of the number of communication rounds (number of phases) multiplied by the number of clients $M$, which is proved to be logarithmic with respect to $T$. However, the amount of information communicated could be non-logarithmic with respect to $T$, because the number of near-optimal nodes could increase exponentially.  In the case when $d = 0$, the amount of information transferred is obviously $\widetilde{\mathcal{O}}(M \log T)$ because the number of near-optimal nodes is bounded by a constant. However, when $d > 0$,  Again, for any fixed $P$ number of phases, the amount of information communicated is $M \sum_{p=1}^P |\mathcal{K}^p|$, which is maximized when $|\mathcal{K}^p|$ increases exponentially. Therefore, again let $h_M = O(\log M)$ be the largest layer such that $\tau_h \leq M$ as in Lemma \ref{lem: length_of_phase}, then before $h_M$, each node will only be pulled once by each client. 
\begin{equation}
\begin{aligned}
    T 
    &\geq \sum_{h=1}^{h_M} k \rho^{-d(h-1)} + \sum_{h=h_M + 1}^P k\rho^{-d(h-1)} \left \lceil \frac{\tau_{h}}{M}  \right  \rceil \\
     &\geq T_0 +\sum_{h=h_M + 1}^P k\rho^{-d(h-1)} \rho^{-2(h-h_M - 1)} \\
    &\geq T_0 + k \rho^{2h_M} \sum_{h=h_M}^{P-1} \rho^{-(2+d)h} = T_0 + \frac{k \rho^{2h_M} }{\rho^{-(2+d)}  - 1} (\rho^{-(2+d)P} - \rho^{-(2+d)h_M})
\end{aligned}
\end{equation}
where $T_0 = \sum_{h=1}^{h_M} k \rho^{-d(h-1)}$ is a constant that only depends on $M$. In order words when $T$ is large enough, $\rho^{-P}  = \mathcal{O}(T^{\frac{1}{2+d}})$. Therefore, the amount of information transferred is of order
\begin{equation}
    M \sum_{h=1}^P \rho^{-d(h-1)} = M (\rho^{-dP} - 1) = \mathcal{O}(M T^\frac{d}{d+2})
\end{equation}

Note that such a communication cost is unfortunately unavoidable by any algorithm. First of all, the lineare dependence on $M$ is unavoidable since we have to communicate with all the clients for synchronous federated learning. For the dependence on $T$, as in the proof of Theorem 13 in \citet{bubeck2011X}, we can construct a hard-instance of the  $\mathcal{X}$-armed bandit problem which is equivalent to a multi-armed bandit problem with $\mathcal{O}(T^\frac{d}{d+2})$ arms, i.e., if we define the (global) objective function to be 
\begin{equation}
    f_{x^*, \eta}(x)=\frac{1}{2}+\max \left\{0, \eta-\ell\left(x, x^*\right)\right\}
\end{equation}
defined within $K$ disjoint balls of radius $\eta$ on the domain $\mathcal{X}$ and with the optimum $x^*$. The reward is a bernouli random variable of the function evaluation. The optimum $x^*$ can be chosen from the center of any of the $K$ balls inside the domain. The problem is proven to be equivalent to a $K$-armed bandit problem with $\eta$ suboptimality gap. As shown in the proof of Theorem 13 in \citet{bubeck2011X}, $K$ is chosen to be $\lceil a_0 \eta^{-d}\rceil$ and $\eta = a_1 T^{\frac{1}{d+2}}$. And thus $K = {\Omega}(T^\frac{d}{d+2})$. We recommend the readers to read the details of the proof. Now since any of the center of the disjoint balls, or equivalently, any arm in the multi-armed bandit problem could be the optimum, any client would have to communicate the reward information of all of them to the server if the algorithm wants to obtain the same regret rate as in our Theorem \ref{thm: regret_upper_bound}. Therefore, such a communication cost is unavoidable.

Nevertheless, as mentioned by \citet{bubeck2011X, Valko13Stochastic, bartlett2019simple}, most functions in real life satisfies that the near-optimality dimension $d = 0$, which means the communicated information is still logarithmic in $T$. In Appendix \ref{app: communication_information_experiments}, we show that this is the case in all of our experiments.

\section{Differentially Private Federated Phased Node Elimination}
\label{app: DP-Fed-PNE}

In this section, we introduce the Differentially Private-Federated-Node Elimination (\texttt{DP-Fed-PNE}) algorithm, which guarantees federated differential privacy as defined in Definition \ref{def: federated_differential_privacy} as well as similar regret bounds in $\mathcal{X}$-armed bandit problems.

\subsection{Federated Privacy Background}

We first recall the definition of $(\epsilon, \delta)$-differential privacy (Definition \ref{def: differential_privacy}) in single-client bandit algorithms as in \citet{shariff2018Differentially}. Although the definition was originally proposed with respect to a multi-armed bandit problem with finite number of arms, it would be straight-forward to extend the definition to the general $\mathcal{X}$-armed bandit case.
\begin{definition}
\label{def: differential_privacy}
\textbf{ \upshape $((\epsilon, \delta)$-differentially private bandit algorithm, \citet{shariff2018Differentially})}. A bandit algorithm $\pi$ is $(\epsilon, \delta)$-differentially private if for any $t >0$ and for all sequences of rewards $r_{1: t}$ and $r_{1: t}^{\prime}$ that differs in at most one time step, we have for any subset $S_{>t} \subseteq \mathcal{X}^{T-t} := \underbrace{\mathcal{X} \times \mathcal{X} \cdots \mathcal{X}}_{T-t} $  of sequence of actions from time $t+1$ to the end of sequence $T$, it holds that:
\begin{equation}
    \mathbb{P}\left(\pi( x_{1: t}, r_{1: t}) \in S_{>t} \right) \leq     \mathbb{P}\left(\pi( x_{1: t}, r_{1: t}^{\prime}) \in S_{>t} \right) e^\epsilon+\delta
\end{equation}
where $\mathcal{X}$ is the set of actions. When $\delta=0$, the algorithm is said to be $\epsilon$-differential private.
\end{definition}
Next, we extend the $(\epsilon, \delta)$-differential privacy in single-client bandit algorithms to the federated case, similar to what \citet{dubey2020differentially} extended from the contextual multi-armed bandit case \citep{shariff2018Differentially}. We present the definition of $(\epsilon, \delta, M)$-differential privacy in $\mathcal{X}$-armed bandit as follows.

\begin{definition}
\label{def: federated_differential_privacy}
In a federated learning setting with $M \geq 2$ clients, a multi-client bandit algorithm $\Pi=\left\{\pi_i\right\}_{i=1}^M$ is $(\epsilon, \delta, M)$-federated differentially private if for any $i, j$ s.t. $i \neq j$, any $t \geq 1$, and set of sequences $R_{t}=\left\{r_{m, 1:t}\right\}_{m=1}^M$ and $R_{t}^{(i)}=\left\{r_{m, 1:t}\right\}_{m=1, m \neq i}^M \cup r_{i, 1:t}^{\prime}$ where $r_{i, 1:t}$ and $r_{i, 1:t}^{\prime}$ are different in at most one time step, and for any subset $S_{>t} \subseteq \mathcal{X}^{T-t} := \underbrace{\mathcal{X} \times \mathcal{X} \cdots \mathcal{X}}_{T-t} $  of sequence of actions from time $t+1$ to the end of sequence $T$, it holds that:

\begin{equation}
    \mathbb{P}\left(\pi_j( x_{j, 1: t}, R_{t}) \in S_{>t} \right) \leq     \mathbb{P}\left(\pi_j( x_{j, 1: t}, R_{t}^{(i)}) \in S_{>t} \right) e^\epsilon+\delta
\end{equation}
\end{definition}

In essence, the above definition means that the reward history of the point (action) chosen by any client at any time should be protected w.r.t. the release of future actions. Note that the above definition does not imply anything about the differential privacy of a single client with respect to its history of actions and rewards. Therefore, we allow the clients to have access to their own past rewards instead of adding  noise to them \citep{Tossou2016Algorithms}. Achieving differential privacy in the single-client $\mathcal{X}$-armed bandit problem could be an interesting future work direction.

\subsection{\texttt{DP-Fed-PNE} Algorithm and Theoretical Analysis}

\begin{algorithm}
   \caption{ \texttt{DP-Fed-PNE: $m$-th client}}
   \label{alg: DP-Fed-PNE_client}
\begin{algorithmic}[1]
   \STATE \textbf{Input:} $k$-nary partition $\mathcal{P}$
   \STATE \textbf{Initialize} $p = 1$
   \WHILE{not reaching the time horizon $T$}
    \STATE Update $p = p+1$
    \STATE  Receive  $\{\mathcal{P}_{h, i}, t_{m, h, i}\}_{(h, i) \in \mathcal{K}^p}$ from the server
    \FOR{ $\mathcal{P}_{h, i}$ with $(h, i) \in \mathcal{K}^p$  }
    \STATE Pull the node for $ t_{m, h, i}$ times, receive rewards $\{r_{m, h, i, t}\}_{t=1}^{t_{m, h, i}}$
    \STATE Add independent Gaussian noise to each reward $\overline{r}_{m,h,i,t} = r_{m,h,i,t} + \mathcal{N}(0, \sigma^2)$
    \STATE Calculate the local mean estimate $\widehat{\mu}_{m, h,i} = \frac{1}{ t_{m, h, i}}  \sum_{t=1}^{t_{m,h,i}} \overline{r}_{m,h,i,t} $
    \ENDFOR
    \STATE Send the local estimates $\{\widehat{\mu}_{m, h,i}\}_{(h,i) \in \mathcal{K}^p}$ to the server
   \ENDWHILE
\end{algorithmic}
\end{algorithm}

\begin{algorithm}
   \caption{ \texttt{DP-Fed-PNE: server}}
   \label{alg: DP-Fed-PNE_server}
\begin{algorithmic}[1]
   \STATE \textbf{Input:} $k$-nary partition $\mathcal{P}$, smooth parameters $\nu_1, \rho$
   \STATE \textbf{Initialize} $ \mathcal{K}^1 = \{(0, 1)\}, h = 0, p = 0$
   \WHILE{not reaching the time horizon $T$}
   \STATE $p = p + 1; h = h+1$
    \WHILE{$|\mathcal{K}^p| \tau_{h}\leq M$ or $\tau_h \leq 1$}
    \STATE $\mathcal{K}^{p} = \left\{(h'+1, ki-j) \mid \forall (h',i) \in \mathcal{K}^p, j \in [k-1] \right\}$ \\  Renew  $h = h+1$ 
    \ENDWHILE
    \STATE Compute the number  $ t_{m,h,i} =  \left \lceil  \frac{\tau_{h}}{M} \right \rceil $ and the phase length $|\mathcal{T}^p| = |\mathcal{K}^p| t_{m, h, i}$ 
    \STATE  Broadcast the set of nodes and pulled times $\{\mathcal{P}_{h, i}, t_{m, h, i}\}_{(h, i) \in \mathcal{K}^p}$ to every client $m$
    \STATE  Receive local estimates $\{\widehat{\mu}_{m, h,i}\}_{m \in [M], (h,i) \in \mathcal{K}^p}$ from the clients
    \FOR{ every $(h, i) \in \mathcal{K}^p$  }
    \STATE Calculate the global mean estimate $\widehat{\mu}_{h,i} = \frac{1}{M} \sum_{m=1}^M \widehat{\mu}_{m, h,i}$
    \ENDFOR
    \STATE Compute $(h^p,i^p) = \arg\max_{(h, i) \in \mathcal{K}^p}\widehat{\mu}_{h,i} $
    \STATE Compute the elimination set \\
    $\mathcal{E}^p = \left \{(h, i) \in \mathcal{K}^p \mid \widehat{\mu}_{h,i} + c \sqrt{\frac{\log(c_1 T/\delta)}{T_{h,i}}}  + \nu_1\rho^h < \widehat{\mu}_{h^p,i^p} - c \sqrt{\frac{\log(c_1 T/\delta)}{T_{h^p,i^p}}}  \right \}$  
    \STATE Compute the new set of nodes $ \mathcal{K}^{p+1} = \left\{(h+1,ki-j) \mid (h,i) \in (\mathcal{K}^p \setminus \mathcal{E}^p), j \in [k-1] \right\}$
   \ENDWHILE
\end{algorithmic}
\end{algorithm}

Note that in the original \texttt{Fed-PNE} algorithm (Algorithms \ref{alg: client} and \ref{alg: server}), we only communicate in a few rounds (logarithmic in terms of the total number of rounds), and we only send out the average rewards from the clients to the server, by Lemma \ref{lem: dp_infers_feddp}. Therefore, we know that as long as the local averages $\widehat{\mu}_{m,h,i}$ are differentially private with respect to the local rewards $\{r_{m,h,i,t}\}$, federated differential privacy is then achieved. Moreover, since we will never come back to a node for its reward statistics after hierarchical eliminations, it means that only one query is made for any node, and thus simply adding one noise would be sufficient.

We thus propose the \texttt{DP-Fed-PNE} algorithm, as shown in Algorithm \ref{alg: client} and \ref{alg: server}. Now compared with the original ones, only a single line of modifications is needed, which is line 8 where we add independent Gaussian noise to each reward (Laplacian noise would be similar). We can set the variance $\sigma^2$ to be
$
    \sigma^2 = \frac{2\log(1.25/\delta)}{\epsilon^2}
$
.

Then we know that the algorithm is $(\epsilon, \delta, M)$ federated differentially private, which we state in Theorem \ref{thm: differential_privacy}.

\begin{theorem}
\label{thm: differential_privacy}
Let $ \sigma^2 = \frac{2\log(1.25/\delta)}{\epsilon^2}
$ in the \texttt{DP-Fed-PNE} algorithm (Algorithm \ref{alg: DP-Fed-PNE_client} and \ref{alg: DP-Fed-PNE_server}), then it is $(\epsilon, \delta, M)$-federated differentially private. 
\end{theorem}

\begin{remark}
The proof of the above theorem follows directly from the work by \citet{dwork2010differential} because the average reward information of each node is only accessed once during the whole optimization process. After each node of every client is pulled and the average reward is transferred, we either expand the node into several children (the reward information is not inherited), or we eliminate the node completely. That means, we never query the reward information of any node more than once.  Note that the sensitivity of the sum is $1$ because the rewards are bounded by $[0, 1]$. Therefore, in Algorithm \ref{alg: DP-Fed-PNE_client} we have added more-than-enough noise ($\mathcal{O}(t_{m,h,i})$ times) to each node. However, since the number of times each node of each client is pulled can be as small as 1, our choice of the noise always guarantees $(\epsilon, \delta, M)$-differential privacy.
\end{remark}

Now we provide the cumulative regret upper bound of \texttt{DP-Fed-PNE} as follows.

\begin{theorem}
\label{thm: DP-FED-PNE_regret_bound}
Suppose that $f(x)$ satisfies Assumption \ref{assumption: local_smoothness}, and $d$ is the near-optimality dimension of the global objective $f$ as defined in Assumption \ref{assumption: near-optimality dimension}. Setting $\delta = 1/M$, $\sigma^2 = \frac{2\log(1.25/\delta)}{\epsilon^2}$, we have the following upper bound on the expected cumulative regret of the \texttt{DP-Fed-PNE} algorithm.
\begin{equation}
\begin{aligned}
\nonumber
  \mathbb{E}[R (T)] 
 \leq C_3(\epsilon, \delta) M^{1-\frac{1}{2\log_k \rho}}  + C_4(\epsilon, \delta) \left( M^{\frac{d+1}{d+2}}  T^{\frac{d+1}{d+2}} (\log (MT))^{\frac{1}{d+2}} \right)\\
\end{aligned}
\end{equation}
where $C_3(\epsilon, \delta)$ and $C_4(\epsilon, \delta)$ are two constants that depend on $(\epsilon, \delta)$
\begin{equation}
\begin{aligned}
     &C_3(\epsilon, \delta) = \frac{48 k \nu_1 }{\rho - \rho^3} \left(\frac{\nu_1^2 \epsilon^2}{ 4\epsilon^2 + 32\log(1.25/\delta)} \right)^{-\frac{1}{2\log_k \rho}} + 1\\
     &C_4(\epsilon, \delta) = 2 \left(\frac{48 k C (4\epsilon^2 +32 \log(1.25/\delta))\rho^{d - 1} (12\nu_1)^{d+1}}{ \nu_1 \epsilon^2 (\rho^{-({d}+1)} - 1)} \right)^{\frac{1}{d+2}} \left(1+ \frac{1}{8} \log (2M)\right).
     \end{aligned}
\end{equation}
\end{theorem}

\begin{remark}
The proof of Theorem \ref{thm: DP-FED-PNE_regret_bound} follows the proof of Theorem \ref{thm: regret_upper_bound}. Since we add Gaussian noise to the rewards, we make the reward distribution $(\frac{1}{4} + \sigma^2)$-sub-Gaussian instead of just bounded. Therefore, we need to change the concentration inequality (Lemma \ref{lem: good_event}) into Lemma \ref{lem: DP_good_event}. As we can observe in the lemma, we only have to change the value $c$ in our proofs, and naturally the threshold value $\tau_h$ for each layer. However, the other lemmas remain the same. Simply substitute the new values of $c, c_1$, and $\sigma^2$ into the final bound in Theorem \ref{thm: regret_upper_bound} would suffice.
\end{remark}

\begin{remark}
The term $C_3(\epsilon, \delta)$ would become smaller when $(\epsilon, \delta)$ are smaller, which is expected since larger noise would be added to the rewards, and thus the threshold becomes larger. Therefore, the waste of samples in the first few phases is smaller and thus the first term is reduced. The second term, becomes naturally larger then $(\epsilon, \delta)$ is smaller, showing higher regret (less utility) when we want stronger privacy.
\end{remark}

\subsection{Useful Lemmas and Definitions}

\begin{lemma}
\label{lem: dp_infers_feddp}
\textbf{\upshape (Proposition 3 in \citet{dubey2020differentially})}
 Consider $P \leq T$ synchronization rounds in the \texttt{DP-Fed-PNE} Algorithm, if the sequence $ \left\{\left \{\widehat{\mu}_{m, h, i}\right\}_{(h,i) \in \mathcal{K}^p} \right \}_{p =1}^P$ is $(\varepsilon, \delta)$-differentially private with respect to $\left\{\left\{r_{m,h,i, t}\right\}_{(h,i) \in \mathcal{K}^p}\right\}_{p =1}^P$, for each client $i \in[M]$, then all clients are $(\varepsilon, \delta, M)$-federated differentially private.
\end{lemma}



 \begin{definition}
 \textbf{(\upshape sub-Gaussian Random Variables)}
 A zero-mean random variable $X \in \mathbb{R}$ is said to be sub-Gaussian with variance proxy $\sigma^2$ if $\mathbb{E}[X]=0$ and its moment generating function satisfies
\begin{equation}
\mathbb{E}[\exp (s X)] \leq \exp \left(\frac{\sigma^2 s^2}{2}\right), \quad \forall s \in \mathbb{R} .
\end{equation}
\end{definition}

\begin{lemma}
  \textbf{\upshape(Concentration of sub-Gaussian Random Variables)}
  \label{lem: concentration_subGaussian}
If $\{X_1, X_2, \cdots, X_n\}$  are independent $\{\sigma_1^2,\sigma_2^2, \cdots, \sigma_n^2 \}$-sub-Gaussian random variables,
\begin{equation}
\mathbb{P}\left(\frac{1}{n} \sum_{i=1}^n\left(X_i-\mathbb{E}\left[X_i\right]\right) \geq t\right) \leq \exp \left(-\frac{n t^2}{\frac{2}{n} \sum_{i=1}^n \sigma_i^2}\right) .
\end{equation}
 \end{lemma}

\begin{lemma}
    \label{lem: DP_good_event}
    \textbf{\upshape (High Probability Event)} Define the ``good" event $E_t$ as
\begin{equation}
    E_t = \left\{ \forall p \in P_t, \forall (h,i) \in \mathcal{K}^p, \forall T_{h,i} \in [MT], | f(x_{h,i}) - \widehat{\mu}_{h,i}| \leq  c \sqrt{\frac{\log(c_1 T/\delta)}{T_{h,i}}} \right\}
\end{equation}
 where the right hand side is exactly the confidence bound  $b_{h,i}$ for the node $\mathcal{P}_{h,i}$ and $c = \sqrt{4+16\sigma^2}, c_1 = (2M)^{1/8}$ are two constants. Then for any fixed round $t$, we have
$
     \mathbb{P}(E_t) \geq 1 - \delta/T^6
$
\end{lemma}
\textbf{Proof}. First of all, note that each $r_{m,h,i,t}$ is $\frac{1}{4}$-sub-Gaussian because it is bounded by $[0, 1]$. Moreover, the added noise is $\mathcal{N}(0, \sigma^2)$ and thus $\sigma^2$-sub-Gaussian. Therefore, the noisy reward $\overline{r}_{m,h,i,t}$ is $(\frac{1}{4}+ \sigma^2)$-sub-Gaussian. For every $p \in P_t$ and every $(h,i) \in \mathcal{K}^p$, note that the node $\mathcal{P}_{h,i}$ is sampled the same number of times $(t_{m,h,i})$ independently from every local objective $m$, therefore by the concentration inequality for sub-Gaussain random variables (Lemma \ref{lem: concentration_subGaussian}), for any $x>0$, we have
\begin{equation}
  \mathbb{P}\left( \frac{1}{T_{h,i}} \left| \sum_{m \in [M]} \sum_{t \in [t_{m,h,i}]} r_{m, t} -  \sum_{m \in [M]} t_{m,h,i} f_m (x_{h,i}) \right| \geq x \right) \leq 2 \exp \left(-\frac{ T_{h,i}x^{2}}{2(\frac{1}{4}+ \sigma^2)}\right).
\end{equation}
which is equivalent as 
\begin{equation}
  \mathbb{P}\bigg( \left | f(x_{h,i}) - \widehat{\mu}_{h,i} \right| \geq x \bigg) \leq 2 \exp \left(-\frac{ T_{h,i}x^{2}}{2(\frac{1}{4}+ \sigma^2)}\right).
\end{equation}
Therefore by the union bound, the probability of the event ${E}_t^c$ can be bounded as
\begin{equation}
\begin{aligned}
\mathbb{P} \left ({E}_{t}^{\mathrm{c}}\right)
&\leq \sum_{p \in P_t } \sum_{(h, i) \in \mathcal{K}^p} \sum_{T_{h,i}=1}^{MT} \mathbb{P}\bigg(| f(x_{h,i}) - \widehat{\mu}_{h,i}| >  b_{h,i} \bigg)\\
&\leq \sum_{p \in P_t } \sum_{(h, i) \in \mathcal{K}^p} 2MT \exp \bigg(- \frac{T_{h,i} b_{h,i}^{2}}{2(\frac{1}{4}+ \sigma^2)} \bigg)\\
& = 2MT \exp \bigg(-\frac{c^2}{2(\frac{1}{4}+ \sigma^2)} \log(c_1 T/\delta) \bigg) \left ( \sum_{p \in P_t} |\mathcal{K}^p| \right) \\
&\leq 2 MT^2 \left(\frac{\delta}{c_1 T} \right)^{\frac{c^2}{2(\frac{1}{4}+ \sigma^2)}} \leq \frac{\delta}{T^6}.
\end{aligned}
\end{equation}

  \hfill $\square$

\section{Experiment Details}
\label{app: experiment}

In this section, we provide the experimental details related to Section \ref{sec: experiments} and the additional experiments.

\subsection{Algorithms and Hyperparameters}

\begin{remark}
Apart from the experimental results on \texttt{HCT}, \texttt{Fed1-UCB}, and \texttt{Fed-PNE}, \texttt{FN-UCB} in Section \ref{sec: experiments}, we have also compared our algorithm with Bayesian Optimization algorithms (\texttt{BO} and \texttt{BBO}). However, these comparisons are unfair because both \texttt{BO} and \texttt{BBO} take a long time to run, due to the high computational complexity of solving Gaussian Process. Also, \texttt{BO} and \texttt{BBO} assume that there is only one objective function, whereas heterogeneous local objectives exist in our problem. Moreover, the performance of these algorithms are not significantly better than our algorithm, even if we give them the advantage to run in their own setting. For example, we allow instantaneous communications in \texttt{BO}. Therefore, we have only conducted experiments on the synthetic functions (Garland and DoubleSine). The results are shown in Figure \ref{fig: All_algos}.  
\end{remark}

\subsubsection{The \texttt{HCT} Algorithm}
The \texttt{HCT} algorithm is originally proposed by \citet{azar2014online} for $\mathcal{X}$-armed bandit problems. For all the experiments, we use a random binary partition of the parameter space, which uniformly randomly selects a dimension of the space when expanding a parent node, and then splits the node into two equal-sized children. The \texttt{HCT} receives rewards directly from the global objective (because there are no clients in the algorithm) while the others receive rewards from the local objective. In such a way of comparison, we show that \texttt{Fed-PNE} is able to outperform \texttt{HCT} even if \texttt{HCT} has the advantage of accessing the global objective. For the two parameters $\nu_1$ and $\rho$, we have followed \citet{li2021optimumstatistical} and used the best parameter settings $\nu_1 = 1$ and $\rho = 0.75$. We have used the publicly-available implementation by \citet{Li2023PyXAB} at the link \url{https://github.com/WilliamLwj/PyXAB}

\subsubsection{The \texttt{KernelTS} Algorithm}

The \texttt{KernelTS} algorithm is proposed by \citet{chowdhury2017kernelized} for kernelized bandit. Similar to \texttt{HCT}, we have run the algorithm on the global objective directly. We have used the publicly available implementation with the default hyperparameter settings at the link \url{https://github.com/ZeroWeight/NeuralTS}.

\subsubsection{The \texttt{Fed1-UCB} Algorithm}
 The \texttt{Fed1-UCB} algorithm is proposed by \citet{shi2021federateda} for the case when the number of clients is fixed in multi-armed bandit problems. Note that for such algorithms to work in $\mathcal{X}$-armed bandit problems, we have to manually choose a set of points inside the parameter space to be the arms. However, such algorithms face the following dilemma when selecting the arms. If too many arms are selected, then the exploration phase length might be very long and the algorithm does not have the chance to perform exploitation. If too few arms are selected, then the reward of selected arms could be very different from the optimum. Specifically in terms of \texttt{Fed1-UCB}, this dilemma comes from the $f(p)$ number of times that each arm has to be pulled \citet{shi2021federateda}. The situation worsens when the dimension of the parameter space increases, resulting in exponentially more arms to choose from. 

\begin{figure*}
\centering
\hspace*{-1.7em}
\subfigure[\footnotesize Regret on Garland]{
  \centering
  \includegraphics[width=0.44\linewidth]{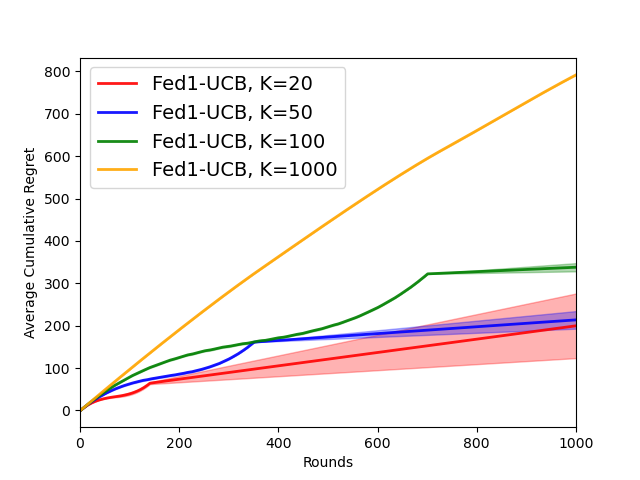}
  \label{fig: FedUCB_Garland_K}
}
\subfigure[\footnotesize Regret on DoubleSine]{
  \centering
  \includegraphics[width=0.44\linewidth]{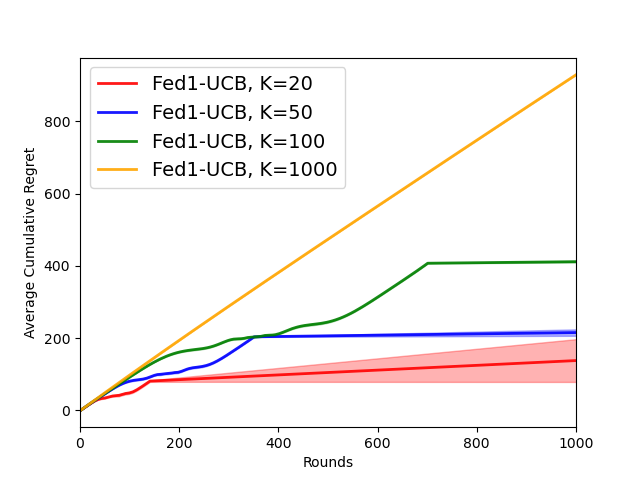}
    \label{fig: FedUCB_DoubleSine_K}
}\hspace*{-1.7em}%
\caption{Average cumulative regret of \texttt{Fed1-UCB} with different number of arms $K$ in the meshgrid.}
\label{fig: FedUCB_K}
\end{figure*}

We have followed the open-source implementation of \texttt{Fed1-UCB} at \url{https://github.com/ShenGroup/FMAB}. However, we have set the tuning parameter $\kappa$ in $f(p)$ to be 1 in our experiments, instead of 10 as in the original repository because \texttt{Fed1-UCB} would not even finish the first phase in our experiments when setting $\kappa = 10$, which means that no arm elimination can be performed. We select the $K$ arms from \textbf{each dimension} of the domain by first randomly selecting a starting point around the boundaries, and then uniformly construct a meshgrid. For example, for the case $\mathcal{X} = [0, 1]$ and $K = 100$, we first select the starting point uniformly randomly from $[0, 0.01]$, and then select the other arms by constructing a uniform mesh from $[0.01, 1]$. The randomization is intended to avoid consistent selection of the same (bad) arms from the domain since there is no prior knowledge about the objective functions. We show the average cumulative regret of \texttt{Fed1-UCB} with different choices of $K$ in Figure \ref{fig: FedUCB_K}. The curves correspond nicely to our description of the dilemma.

\subsubsection{The \texttt{FN-UCB} algorithm}

The \texttt{FN-UCB} algorithm is proposed by \citet{dai2023federated} to utilize neural networks in the federated multi-armed bandit problem. Similar to \texttt{Fed1-UCB}, the algorithm only works when we manually choose a set of points inside the parameter space to be the $K$ arms observed in each round. Therefore, we have again generated the arms by constructing a meshgrid on the parameter domain, as we did in the implementation for the \texttt{Fed1-UCB} algorithm. We have followed the implementation by the authors \citet{dai2023federated} at the link \url{https://github.com/daizhongxiang/Federated-Neural-Bandits}. It's worth mentioning that \texttt{FN-UCB} takes very long (20x more than \texttt{Fed-PNE}) to optimize the objectives because the algorithm needs to train neural networks.

\subsubsection{The \texttt{Fed-PNE} Algorithm}
Similar to \texttt{HCT}, for all the experiments, we use a random binary partition of the parameter space. For the hyper-parameters of our algorithm, we use the following values. 
\begin{itemize}
    \item The smoothness parameters $\nu_1$ and $\rho$ are set to be $\nu_1 = 1$ and $\rho = 0.5$.
    \item The confidence parameters $c$ and $c_1$ are set to be $c = 0.1$ and $c_1 = 1$. 
\end{itemize}
We have also tuned these parameters, such as using $c = 2$ and $c_1 = (2M)^{1/8}$ following Lemma \ref{lem: good_event}. We find that the algorithm's performance is quite robust to different values of $c_1$, but using a large $c$ will make the algorithm explore the parameter space much slower (and thus it generates larger regret). At the same time, the variance of the results becomes much smaller, i.e., the shaded regions in our figures become smaller (or even vanish), which is expected since we have larger confidence on the ``good" event $E_t$. The algorithm behaves quite well with the default $\nu_1$ and $\rho$.

\begin{figure*}
\centering
\hspace*{-1.7em}
\subfigure[\footnotesize Regret on Garland]{
  \centering
  \includegraphics[width=0.44\linewidth]{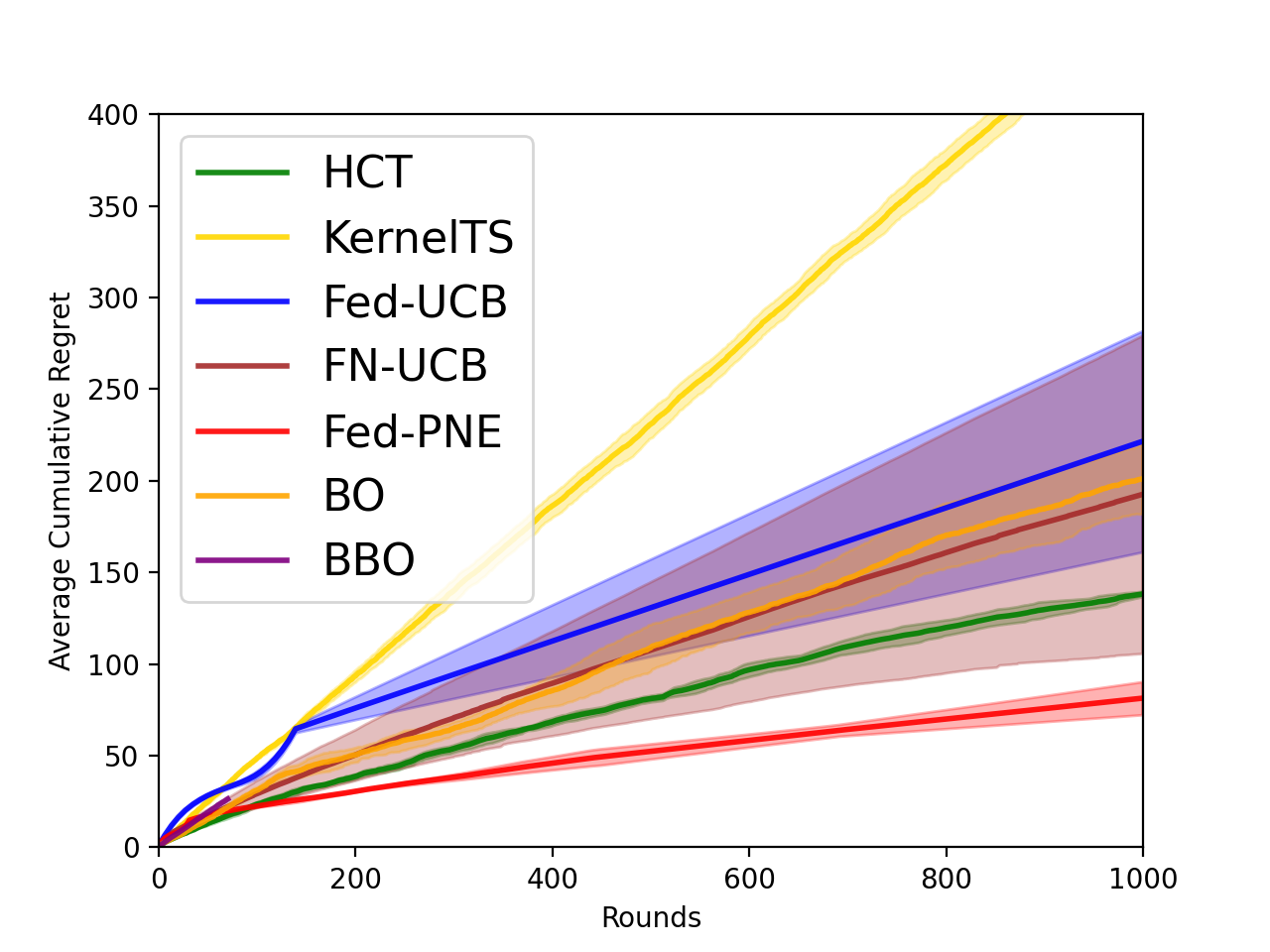}
  \label{fig: All_garland}
}
\subfigure[\footnotesize Regret on DoubleSine]{
  \centering
  \includegraphics[width=0.44\linewidth]{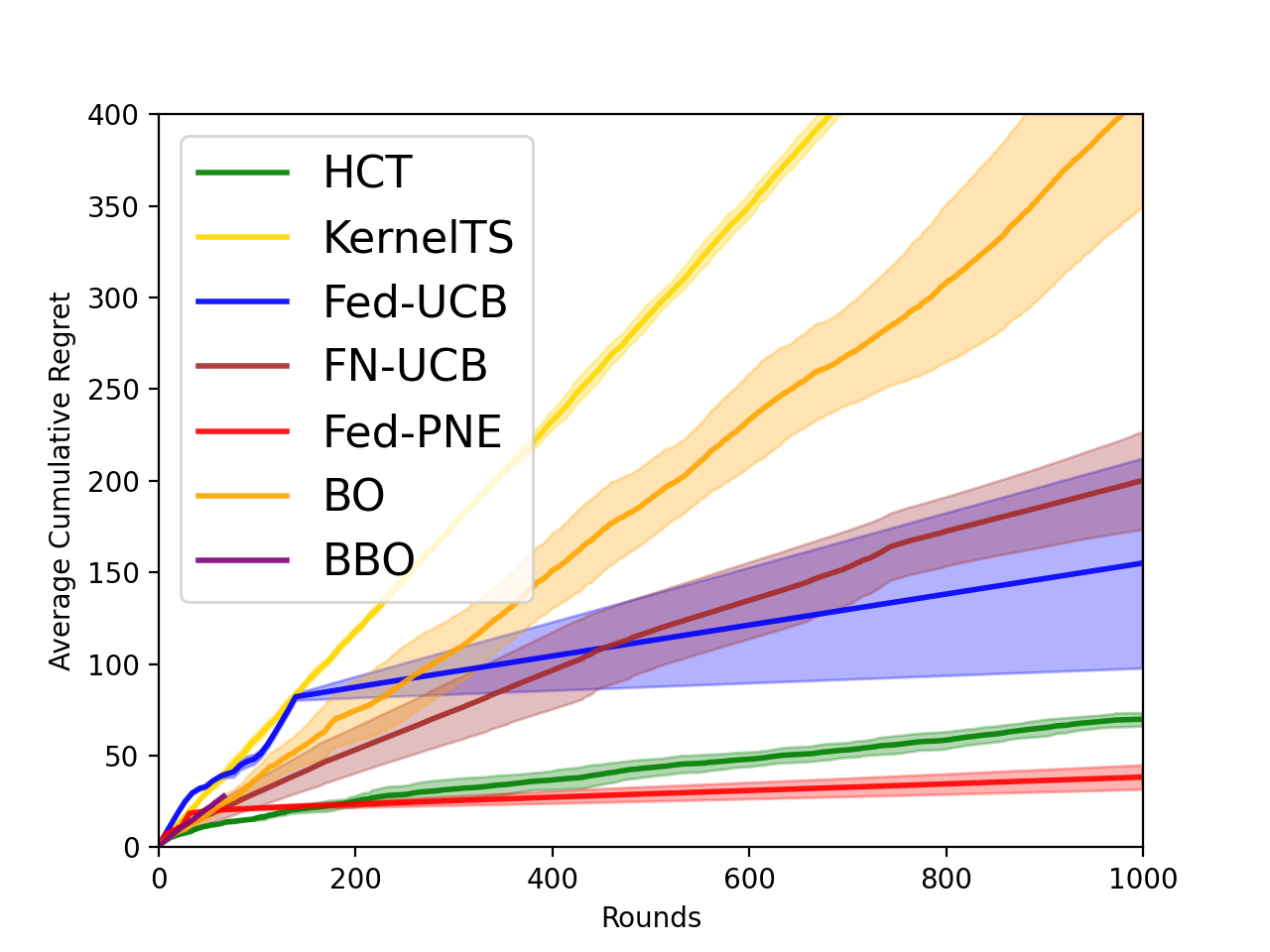}
    \label{fig: All_doublesine}
}\hspace*{-1.7em}%
\caption{Average cumulative regret of different algorithms on the synthetic functions. Unlimited communications are allowed for \texttt{HCT}, \texttt{KernelTS} and \texttt{BO}. \texttt{BBO} takes a whole day to only run less than 100 rounds, whereas the other algorithms take at most a few hours to run 1000 iterations (\texttt{BO}).}
\label{fig: All_algos}
\end{figure*}

\subsubsection{The \texttt{BO} Algorithm}

We have used the public implementation of Bayesian Optimization (\texttt{BO}) at the link \url{https://github.com/SheffieldML/GPyOpt} \citep{frazier2018tutorial}. For all the settings of \texttt{BO} such as the the acquisition function and the prior, we have directly used the default setting. Since \texttt{BO} is a single-client algorithm, we have directly run it on the global objective as we did for \texttt{HCT}. However, \texttt{BO} runs very slowly compared with $\mathcal{X}$-armed bandit algorithms, which is expected since the algorithm involves inverting a matrix of size $n$ by $n$ given $n$ evaluations. Therefore, when the total number of evaluations is large (e.g.), \texttt{BO} takes hours to finish the experiments. Moreover as shown in Figure \ref{fig: All_algos}, the cumulative regret of \texttt{BO} on the two synthetic functions is not comparable with that of \texttt{Fed-PNE}.

\subsubsection{The \texttt{BBO} Algorithm}
There are many Batched Bayesian Optimization (\texttt{BBO}) algorithms, and we have used the one proposed by \citet{wang2018batched} and its open-source implementation at the link \url{https://github.com/zi-w/Ensemble-Bayesian-Optimization}. For the parameter settings, we have used the default implementations. However, \texttt{BBO} algorithms assume that there is only one objective and is not directly applicable to solve the federated problem. Therefore, we have directly treated the function values of the local objectives as if there are from one objective. However, \texttt{BBO} is very slow because it receives $M$ rewards in one round, and thus it needs to solve a matrix inverse problem of size $Mt \times Mt$ at round $t$. Therefore, \texttt{BBO} is even slower than \texttt{BO} and it takes 24 hours to run less than 100 rounds. We are unable to finish the experiments of \texttt{BBO} within a reasonable amount of time and we only show partial results in Figure \ref{fig: All_algos}. Moreover, the cumulative regret of \texttt{BBO} does not show any advantages in the early stage.

\subsection{Datasets and Objectives in Section \ref{sec: experiments}}

\begin{figure*}
\centering
\hspace*{-1.7em}
\subfigure[\footnotesize Garland function]{
  \centering
  \includegraphics[width=0.44\linewidth]{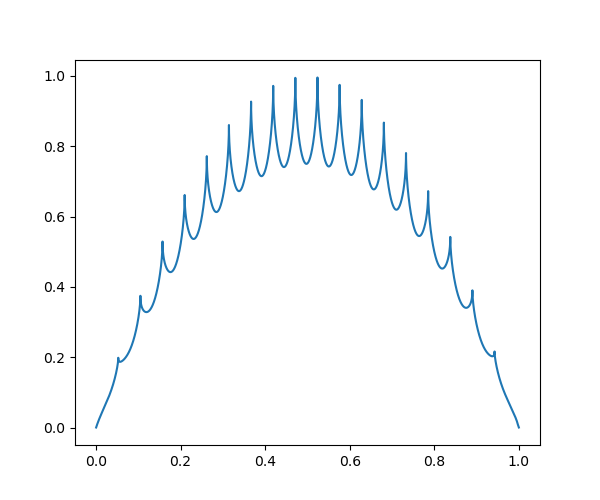}
  \label{fig: Garland_plot}
}
\subfigure[\footnotesize DoubleSine function]{
  \centering
  \includegraphics[width=0.44\linewidth]{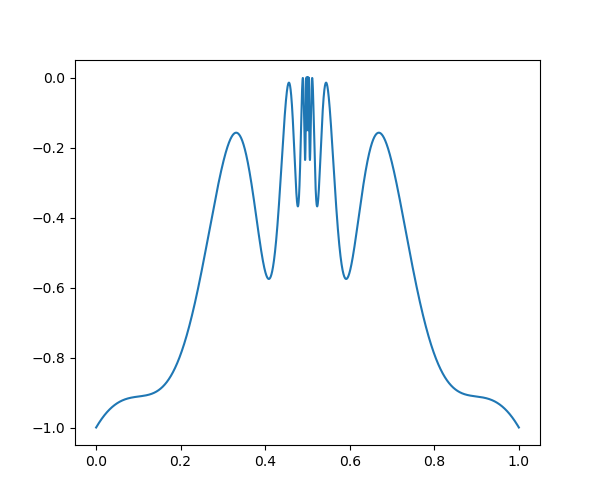}
    \label{fig: DoubleSine_plot}
}\hspace*{-1.7em}%
\caption{Plot of the two synthetic functions used in the experiments}
\label{fig: synthetic_obj}
\end{figure*}

\textbf{Synthetic Functions}. Garland and DoubleSine are two synthetic functions that are used very frequently in the experiments of $\mathcal{X}$-armed bandit algorithms because of their large number of local optimums and their extreme unsmoothness, which appeared in works such as \citet{azar2014online, Grill2015Blackbox, shang2019general, bartlett2019simple, li2021optimumstatistical}. The mathematical expression of these two functions are $f(x) = x(1-x)(4-\sqrt{|\sin(60x)|}$ for Garland and $f(x)=s(\frac{1}{2}\log_2 | 2x-1|)(| 2x-1|^{-\log_2 \rho_2 } - (2x-1)^{-\log_2 \rho_1 }) - (| 2x-1|)^{-\log_2 \rho_1 }  $ for DoubleSine. Both functions are defined on the domain $[0, 1]$. We provide the plots of these functions in Figure \ref{fig: synthetic_obj}. As shown in the two figures, these functions are very hard to optimize. The global objective $f$ is defined to be the original synthetic functions while for the client objectives, we add a (consistent) unit-normal random perturbation to the original functions.

\textbf{Landmine Dataset}.
The landmine dataset consists of 29 landmine fields with features from radar images to detect whether a certain position has landmine or not. The dataset can be downloaded from \url{http://www.ee.duke.edu/~lcarin/ LandmineData.zip}. We have used the publicly available code from \citet{dai2020federated} at \url{https://github.com/daizhongxiang/Federated_Bayesian_Optimization} to process the data. We tune two hyper-parameters of the SVM model, specifically the RBF kernel parameter and the $L_2$ regularization from the domain [[0.01, 10], [1e-4, 10]]. We have followed \citet{dai2020federated} and split the dataset into equal-sized training set and testing set. The SVM is trained on the training set with the selected hyper-parameter and then evaluated on the testing set. Each client chooses one landmine (randomly) and thus only observes local data. We use the AUC-ROC score as the objectives to be optimized. The local objectives are the AUC-ROC scores on one landmine, and the global objective is the AUC-ROC score on average. Note the for the computation of the sample cumulative regret, we assume the unknown global optimal to be 1.

\textbf{COVID-19 Vaccine Dosage Prediction}. 
We use the famous Susceptible-Exposed-Infected-Recover (SEIR) model as the epidemiological model to describe COVID-19, which consists of a system of ordinary differential equations (ODEs) whose solutions are the predicted numbers of people in each state. We have used the open source code by \url{https://github.com/cfculhane/coronaSEIR}, and adding vaccine as a factor that affects the system. (There are many variants of SEIR and we use the model in the aforementioned repository). Specifically, we have the following system of ODEs, 
\begin{equation}
\left \{
\begin{aligned}
        &\frac{dS_t}{d t} = -\frac{ \beta S_t I_t}{N} - \alpha V \\
        &\frac{dE_t}{d t} = \frac{\beta  S_t  I_t}{N} - \sigma E_t \\
        &\frac{dI_t}{d t} = \sigma  E_t - \gamma I_t \\
        &\frac{dR_t}{d t} = \gamma  I_t + \alpha V \\
\end{aligned}
\right.
\end{equation}
where $S_t$ represents the number of people susceptible, $E_t$ represents the number of  people  exposed, $I_t$ represent the number of people infected, and $R_t$ represents the number of people recovered.
In the above equations, $\beta$ is the transmission rate of the disease, $\gamma$ is the recovery rate, $\sigma$ is the propagation rate from exposed to infectious, $\alpha$ is the effective rate of the vaccine and $V$ is the number of people receiving the vaccine.

To tune the best vaccine dosage $x \in [0, 1]$, we have assumed that the number of people getting the vaccine is inversely proportional to the dosage, i.e., $V \propto 1/x$, which is natural since the total volume of the dosages would be the same. For the effective rate, we have assumed a quadratic decrease in the rate when the dosage decreases, and the effective rate decreases to zero when we use only half of the dosages. The local objectives are the final percentages of infected people in one counutry/region and the global objective is the average percentage.
Note the for the computation of the sample cumulative regret, we assume the unknown global optimal to be 0.

\subsection{Benefit of Large Number of Clients}

\begin{figure}[!ht]
\centering
\subfigure[\footnotesize Garland]{
  \centering
  \includegraphics[width=0.44\linewidth]{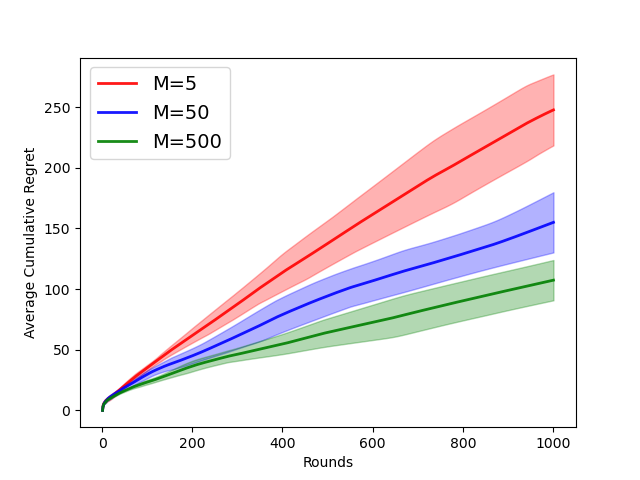}
  \label{fig: Garland_M}
}%
\subfigure[\footnotesize DoubleSine]{
  \centering
  \includegraphics[width=0.44\linewidth]{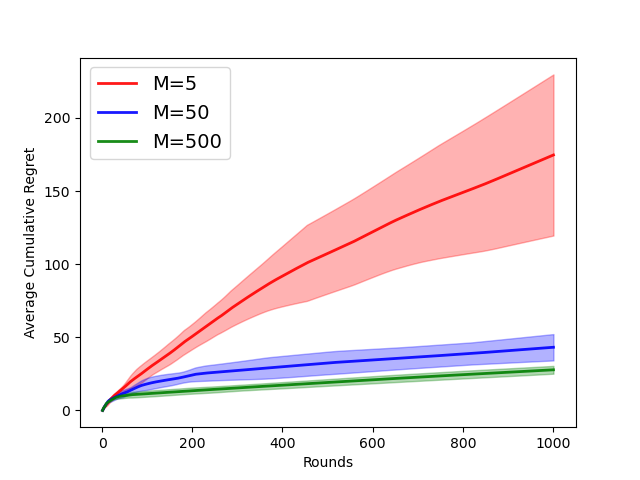}
    \label{fig: DoubleSine_M}
}\hspace*{-1.5em}%
\\
\subfigure[\footnotesize Landmine]{
  \centering
  \includegraphics[width=0.44\linewidth]{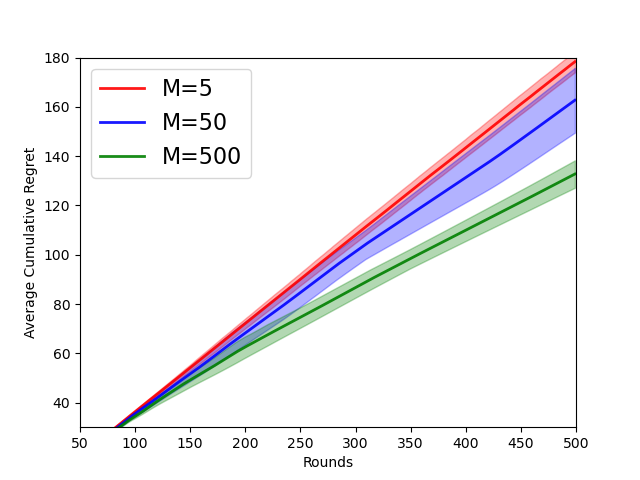}
  \label{fig: Landmine_M}
}%
\subfigure[\footnotesize COVID]{
  \centering
  \includegraphics[width=0.44\linewidth]{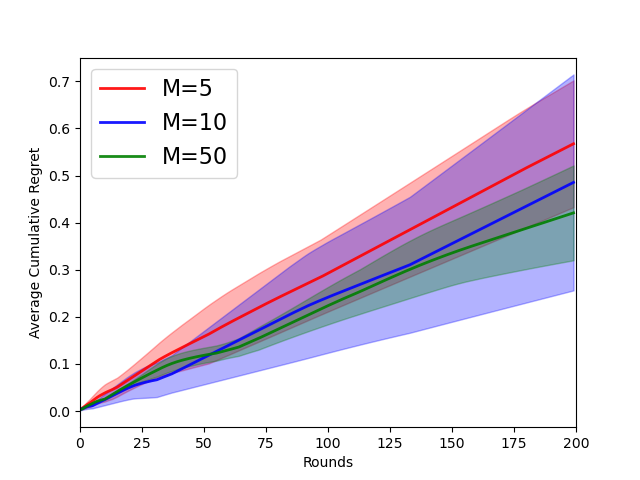}
  \label{fig: COVID_M}
}
\caption{Average cumulative regret of \texttt{Fed-PNE} on the synthetic functions and the real-life datasets with different number of clients. $M$ denotes the number of clients. }
\label{fig: experiments_M}
\end{figure}

We validate the theoretical benefit of having large number of clients when running \texttt{FedPNE}. We plot the average cumulative regret of our algorithm on the four  objectives with different number of clients $M$ in Figure \ref{fig: experiments_M}
As shown in the four figures, larger number of clients improves the average cumulative regret of \texttt{FedPNE}, which proves the correctness of our conclusions in Theorem \ref{thm: regret_upper_bound}.

\subsection{Communication Information}
\label{app: communication_information_experiments}
\begin{figure}[h]
\centering
\hspace*{-1.5em}
\subfigure[\footnotesize Garland]{
  \centering
  \includegraphics[width=0.27\linewidth]{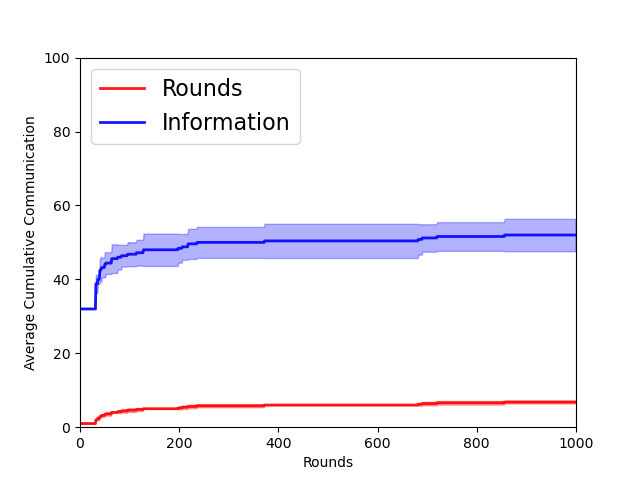}
  \label{fig: Garland_communication}
}\hspace*{-1.5em}%
\subfigure[\footnotesize DoubleSine]{
  \centering
  \includegraphics[width=0.27\linewidth]{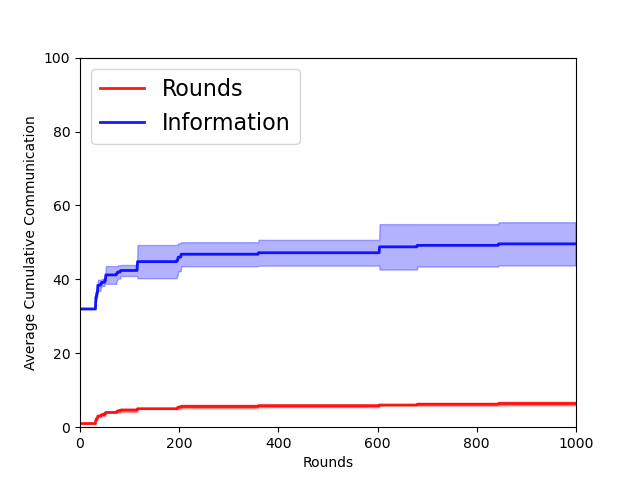}
    \label{fig: DoubleSine_communication}
}\hspace*{-1.5em}%
\subfigure[\footnotesize Landmine]{
  \centering
  \includegraphics[width=0.27\linewidth]{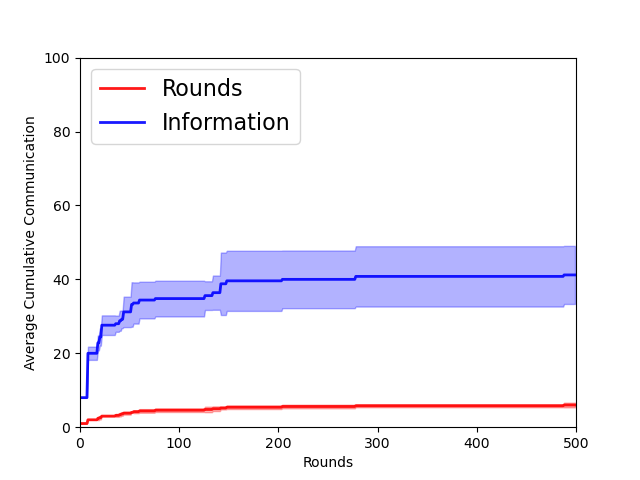}
  \label{fig: Landmine_communication}
}\hspace*{-1.5em}%
\subfigure[\footnotesize COVID]{
  \centering
  \includegraphics[width=0.27\linewidth]{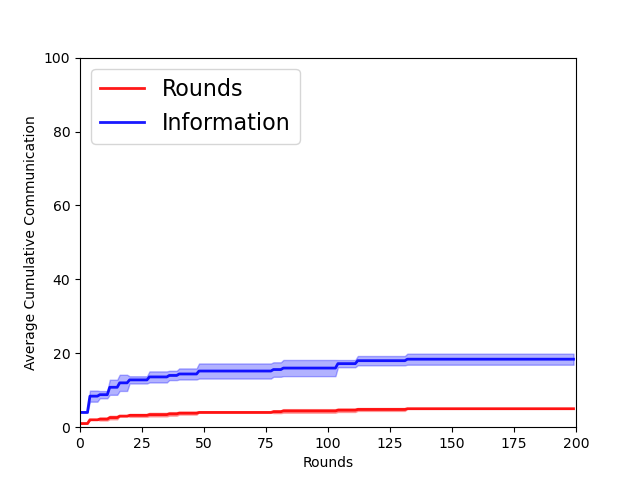}
  \label{fig: COVID_communication}
}
\caption{Number of communication rounds and averaged communicated information of \texttt{Fed-PNE} on the different datasets.}
\label{fig: experiments_communication}
\end{figure}

We have plotted the number of communication rounds and the averaged information (number of nodes) communicated per client in Figure \ref{fig: experiments_communication}. As can be observed, the communication cost in terms of the number of communication rounds is indeed logarithmic with respect to $T$, validating our theoretical claims in Theorem \ref{thm: DP-FED-PNE_regret_bound}. Moreover, the amount of communicated information is logarithmic with respect to $T$ in both the synthetic dataset and the real-life datasets.

\subsection{Additional Experiments of Tuning Neural Networks on MNIST}
\label{subsec: mnist}

\textbf{MNIST Dataset and Neural Network.} The MNIST dataset is one of the most well-known datasets for image classification, which can be downloaded from \url{http://yann.lecun.com/exdb/mnist/}. The dataset consists of 60 thousand images of hand-written digits (i.e., 0-9) to be classified, where 50 thousand are normally the training images and 10 thousand are the testing images. We have utilized the publicly available code at \url{https://github.com/yueqiw/OptML-SVRG-PyTorch} for neural network training on MNIST.  The neural network is a simple two-layer feed-forward network with 64 hidden units. We use the ReLu activation for the hidden layer. For each client, we have trained a neural network using stochastic gradient descent (SGD) with three different hyper-parameters, specifically, the mini batch-size when loading data from [1, 100], the learning rate from [1e-6, 1], and the weight decay from [1e-6, 5e-1].

\textbf{Experiment Setting}. For the local datasets, we randomly sample 500 images from the training set to each client and use them as local training data, and 100 images from the testing set as the local testing data. The reward (and the regret) is defined to be the testing accuracy on the testing set. 
Following the setting in Section \ref{sec: prelim}, to compute sample cumulative regret, one need to have all the testing accuracy of each local data set under all possible hyperparameters settings visited by local search. For example, suppose that we have 10 clients, and thus 10 neural networks have to be trained (till convergence) and tested in each round. Then to compute the regret of this round, one has to additionally train 90 neural networks, leading to a huge computational cost. Therefore, we approximate the regret of each round by the testing accuracy on 1k total testing images when training the network on 5k total training images, thus only 10 additional neural networks training processes are need to compute the regret. Note we also assume the unknown global optimal to be 100\%.

\begin{figure*}
\centering
\hspace*{-1.7em}
\subfigure[\footnotesize Comparison of Regret ($M=10$)]{
  \centering
  \includegraphics[width=0.44\linewidth]{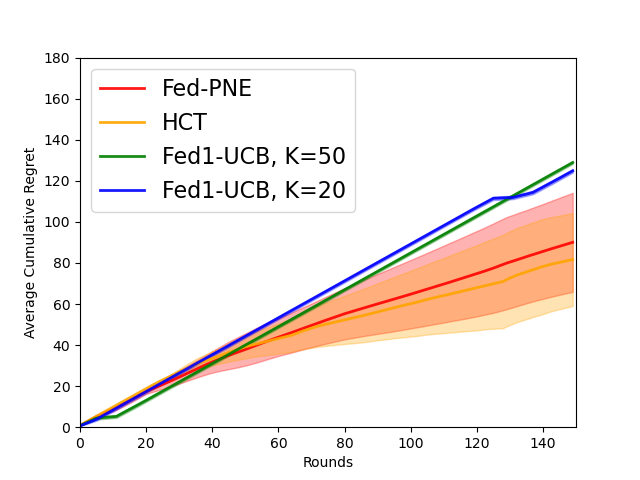}
  \label{fig: MNIST}
}
\subfigure[\footnotesize Regret of Different M]{
  \centering
  \includegraphics[width=0.44\linewidth]{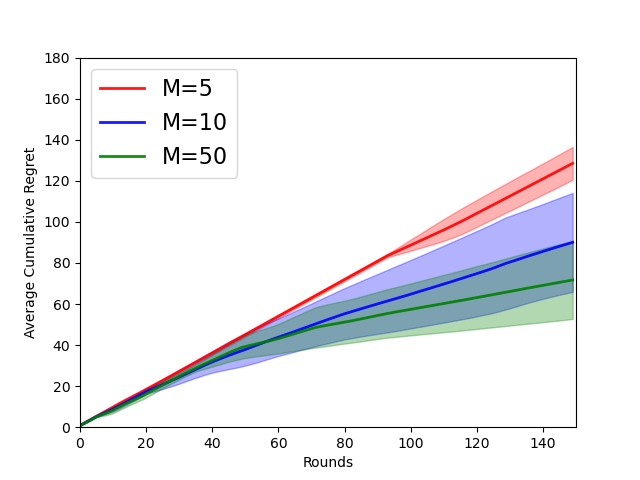}
    \label{fig: MNIST_M}
}\hspace*{-1.7em}%
\caption{(a). Average cumulative regret of different algorithms on MNIST. (b) Average cumulative regret of \texttt{Fed-PNE} on MNIST with different number of clients $M$. }
\label{fig: mnist_results}
\end{figure*}


\textbf{Results and Conclusions.}
We provide the average cumulative regret of the different algorithms when $M=10$ in Figure \ref{fig: MNIST}, and the averaged cumulative regret of \texttt{Fed-PNE} with different $M$ in Figure \ref{fig: MNIST_M}. As can be observed in those figures, similar conclusions as in Section \ref{sec: experiments} about the effectiveness of our algorithm can be made.
We believe that our experimental results already show the advantages of \texttt{FedPNE} over the other algorithms, and the benefit of having many clients.

\end{document}